\documentclass[sigconf]{acmart}
\pdfoutput=1

\AtBeginDocument{%
  \providecommand\BibTeX{{%
    \normalfont B\kern-0.5em{\scshape i\kern-0.25em b}\kern-0.8em\TeX}}}

\copyrightyear{2021}
\acmYear{2021}
\setcopyright{acmlicensed}\acmConference[CHI '21]{CHI Conference on Human Factors in Computing Systems}{May 8--13, 2021}{Yokohama, Japan}
\acmBooktitle{CHI Conference on Human Factors in Computing Systems (CHI '21), May 8--13, 2021, Yokohama, Japan}
\acmPrice{15.00}
\acmDOI{10.1145/3411764.3445296}
\acmISBN{978-1-4503-8096-6/21/05}

\let\citep\cite
\let\citet\cite

\usepackage{balance}
\usepackage{graphics}
\usepackage[T1]{fontenc}
\usepackage{color}
\usepackage{booktabs}
\usepackage{textcomp}
\usepackage{multirow}
\usepackage{ccicons}          
\usepackage[utf8]{inputenc} 
\usepackage{bbm}

\usepackage{amsmath}
\usepackage{amsfonts}
\usepackage{caption}
\usepackage[linesnumbered,lined,ruled,vlined,commentsnumbered]{algorithm2e}
\usepackage{float}

\def \enc{\mathtt{enc}}

\usepackage{tabularx}

\newcommand{\round}[1]{\lfloor#1\rceil}

\def \signifThresh {0.00011}

\newcommand{\appropto}{\mathrel{\vcenter{
  \offinterlineskip\halign{\hfil$##$\cr
    \propto\cr\noalign{\kern2pt}\sim\cr\noalign{\kern-2pt}}}}}

\usepackage{bbding}

\begin{document}

\title{Evaluating the Interpretability of Generative Models by Interactive Reconstruction}

\author{Andrew Slavin Ross}
\affiliation{%
  \institution{Harvard University}
   \city{Cambridge}
  \state{MA}
  \country{USA}
}
\email{andrew\_ross@g.harvard.edu}
\author{Nina Chen}
\affiliation{%
  \institution{Harvard University}
     \city{Cambridge}
  \state{MA}
  \country{USA}
}
\email{nchen1@college.harvard.edu}

\author{Elisa Zhao Hang}
\affiliation{%
  \institution{Harvard University}
     \city{Cambridge}
  \state{MA}
  \country{USA}
}
\email{ezhaohang@college.harvard.edu}

\author{Elena L. Glassman}
\affiliation{%
  \institution{Harvard University}
     \city{Cambridge}
  \state{MA}
  \country{USA}
}
\email{glassman@seas.harvard.edu}

\author{Finale Doshi-Velez}
\affiliation{%
  \institution{Harvard University}
     \city{Cambridge}
  \state{MA}
  \country{USA}
}
\email{finale@seas.harvard.edu}

\begin{abstract}
For machine learning models to be most useful in numerous sociotechnical systems, many have argued that they must be human-interpretable. However, despite increasing interest in interpretability, there remains no firm consensus on how to measure it. This is especially true in representation learning, where interpretability research has focused on ``disentanglement'' measures only applicable to synthetic datasets and not grounded in human factors. We introduce a task to quantify the human-interpretability of generative model representations, where users interactively modify representations to reconstruct target instances. On synthetic datasets, we find performance on this task much more reliably differentiates entangled and disentangled models than baseline approaches. On a real dataset, we find it differentiates between representation learning methods widely believed but never shown to produce more or less interpretable models. In both cases, we ran small-scale think-aloud studies and large-scale experiments on Amazon Mechanical Turk to confirm that our qualitative and quantitative results agreed.
\end{abstract}

\maketitle



\section{Introduction}
Many have speculated that machine learning (ML) could unlock significant new insights in science and medicine thanks to the increasing volume of digital data \cite{hansen2014big}. 
However, much of this data is unlabeled, making it difficult to apply many traditional ML techniques. Generative modeling, a subfield of ML that does not require labels, promises to help by distilling high dimensional input data into lower dimensional meaningful axes of variation, which we call representations. To be most useful in ``unlocking insights,'' though, these representations must be understood by human researchers.

Motivated by the need for human understanding, a burgeoning research area of interpretable ML has emerged~\cite{doshi2017towards,gilpin2018explaining}.  While some of this work has used user studies to quantify interpretability~\cite{huysmans2011empirical,krause2016interacting,poursabzi2018manipulating,lage2019evaluation}, there have been concerns within the HCI community that these studies do not generalize to more realistic use cases \cite{buccinca2020proxy}. These studies are also largely in the context of discriminative rather than generative modeling---even in the few that consider representations \cite{arendt2020parallel}.
Within generative modeling, ML researchers have tried to quantify interpretability via measures of \emph{disentanglement}, which measure how well individual representation dimensions match ground-truth factors that generated the data \cite{ridgeway2016survey}. However, this work is not tested on actual human users, nor are disentanglement measures computable without knowing ground-truth factors. 

In this work, we develop a method for quantifying the interpretability of generative models by measuring how well users can interactively manipulate representations to reconstruct target instances.  
To validate it, we use both MTurk and lab studies to determine whether models known to be understandable a priori can be distinguished from those known to be complex, and also whether our quantitative metrics match qualitative feedback from users.  We also investigate the relationship between our human-grounded interpretability measures and synthetic disentanglement measures.

Our main contributions are as follows: 
\begin{itemize}
\item A task for evaluating the interpretability of generative models, where users interactively manipulate representation dimensions to reconstruct target instances.
\item Large-scale experiments on Amazon Mechanical Turk and smaller-scale think-aloud studies showing our task distinguishes entangled from disentangled models and that performance is meaningfully related to human model understanding, as demonstrated and reported by study participants.
\item Novel results suggesting that ML methods which improve disentanglement on synthetic datasets also improve interpretability on real-world datasets.
\end{itemize}

\section{Related Work}

\subsection{Human-Centered Interpretability Measures}
\label{sec:hci-quant}

While there have been criticisms that ``interpretability'' is ill-defined \cite{lipton2018mythos}, several works have focused on quantifying it, particularly for discriminative models \cite{lim2009and,doshi2017towards,whatcanAIdoforme,abdul2020cogam,lage2019evaluation,allahyari2011user,poursabzi2018manipulating,slack2019assessing,schmid2016does,huysmans2011empirical,interacting-with-predictions}. 
Per Doshi-Velez and Kim~\cite{doshi2017towards} and Miller~\cite{miller2019explanation},
these works typically ground interpretability as the capacity of the model to be sufficiently understood in an appropriate context, and operationalize it as a user's ability to perform various tasks given visualizations of a model according to some performance measures---where the tasks and measures are presumed to be relevant to the desired context. They then study the effect of varying visualizations or models on task performance measures.


%

As a concrete example, Kulesza et al.~\cite{kulesza2013too} present different visualizations of a random forest~\cite{breiman2001random} model, and as their task, ask users a series of questions in a thinkaloud-style protocol~\cite{lewis1982using}. Their performance measure is the difference between the number of accurate and inaccurate statements about the model made by the user, which they compute by transcribing interviews and individually categorizing each statement. As their theoretical grounding, they draw on notions of understanding from Norman~\cite{norman2013design}, who defines understanding in terms of user mental model accuracy. They find that visualizations produce more accurate mental models when complete (the whole truth) and sound (nothing but the truth), even if satisfying those conditions dramatically increases complexity.

More commonly, interpretability is operationalized as simulability: whether humans can use visualizations to predict the behavior of the model in new circumstances~\cite{lim2009and,doshi2017towards, huysmans2011empirical, poursabzi2018manipulating, slack2019assessing, lage2019evaluation, lage2018human}. Though the theoretical grounding of this method is perhaps less clear than the mental model accuracy paradigm of Kulesza et al.~\cite{kulesza2013too} or the cognitive load paradigm of Abdul et al.~\cite{abdul2020cogam}, simulation tasks have the advantage of being model-agnostic and easy to analyze programmatically, which allows them to be used in semi-automated human-in-the-loop optimization procedures such as Lage et al.~\cite{lage2018human}. However, such simulability tasks generally do not present ``the whole truth'' of the model at once.
Our method extends the simulability paradigm from discriminative to generative modeling, but in a way that presents the whole truth of the model.

\subsection{Interpretable Representation Learning}
\label{sec:disentanglement}

\textbf{Background.} Representation learning is generally concerned with finding ways of associating high-dimensional \emph{instances}, which we denote $x$, with low-dimensional \emph{representations}, which we denote $z$. Representation learning methods roughly fall in two categories: \emph{generative modeling}, which maps low-dimensional representations $z$ to high-dimensional instances $x$, and embedding, which maps high-dimensional instances $x$ to low-dimensional representations $z$. Examples of generative models include Hidden Markov Models \cite{ghahramani1996factorial}, Latent Dirichlet Allocation \cite{blei2003latent}, and GANs \cite{goodfellow2014generative}. Examples of embeddings include t-SNE \cite{maaten2008visualizing} and the latent spaces of deep classification models. Examples of both simultaneously (i.e. approximately invertible mappings) include PCA \cite{jolliffe1986principal} and autoencoders \cite{hinton2006reducing}.

\textbf{Disentangled representations and disentanglement measures.} 
Disentangled representations \cite{bengio2013representation,desjardins2012disentangling,higgins2017beta,chen2016infogan} seek mappings between high-dimensional inputs and low-dimensional representations such that representation dimensions correspond to the ground-truth factors that generated the data (which are presumed to be interpretable). 
To evaluate a model's disentanglement, many papers compute \textit{disentanglement measures} with respect to known ground-truth factors.  However, not only do there exist many competing measures \cite{higgins2017beta,chen2018isolating,eastwood2018framework,locatello2018challenging,higgins2018towards,sepliarskaia2019evaluating}, but, to our knowledge, there exists no work that compares them to human notions of understandability.  Our work both performs this comparison and provides a way to evaluate 
representation learning methods on real-world datasets, where we cannot rely on disentanglement measures because the ground truth is generally unavailable to us.

\textbf{Explaining and visualizing representations.} 
To be understood, a representation must be visualized or explained.  
For generative models, this usually means explaining each dimension. Many works \cite{kingma2013auto,higgins2017beta,chen2016infogan,chen2018isolating,kim2018disentangling} show a ``latent traversal'' of instances with linearly varying values of a specific dimension and constant values of other dimensions. Others find or construct ``exemplar'' instances that maximize or minimize particular (combinations) of dimensions~\cite{olah2017feature,olah2018building,alvarez2018towards}.
Another visualization technique, less common because it requires setting up an interactive interface, is to let users dynamically modify representation dimensions and see how corresponding instances change in real time, e.g. using sliders~\cite{ha2018worldmodels}. Related approaches have been explored for discriminative models with predefined meaningful features, e.g. Krause et al.~\cite{interacting-with-predictions}, Wexler et al.~\cite{wexler2019if}, and Cai et al.~\cite{cai2019human}, who use sliders to display exemplars matching user-defined concept activation vectors~\cite{kim2018interpretability}. However, to our awareness, interactive manipulation of autonomously learned generative model dimensions has not been considered in an HCI context, especially to quantify interpretability. We use interactive slider-based visualizations as a subcomponent of our task, and test against baseline approaches using exemplars and traversals.

We do note there are other representation learning visualization methods that could be used in interpretability quantification but do not provide insight into the individual meanings of dimensions and are often specific to embeddings rather than generative models. Examples include Praxis~\cite{cavallo2018visual}, the embedding projector~\cite{smilkov2016embedding}, and a parallel-coordinates-inspired extension by Arendt et al.~\cite{arendt2020parallel}, which further reduce embedding dimensionality down to 2D or 3D with PCA or t-SNE \cite{maaten2008visualizing}. All of these methods are interactive and can help users understand the geometry of the data and what information the representation preserves, but they do not explain the dimensions of variation themselves.  Bau et al.~\cite{bau2017network,bau2019visualizing} both visualize and quantify the interpretability of representations by relating dimensions to dense sets of auxiliary labels, but their approach (while highly effective) is not applicable to most datasets. Our focus is on quantifying how well users can understand representations on their own, in terms of their dimensions, without further projection or side-information.

\section{Approach} \label{sec:method}
We now define our proposed task for evaluating the interpretability of generative models, which we call ``interactive reconstruction,'' and ground it as a measure of understanding.

\subsection{The Interactive Reconstruction Task}
\label{sec:fr-definition}

Assume we are given a generative model $g(z)$: $\mathbb{R}^{D_z} \to \mathbb{R}^{D_x}$,  which maps from representations $z$ to instances $x$. Also assume we are given a distribution $p(Z)$ from which we can sample representations $z$ (which may be approximated by sampling from a dataset $\{z^{(1)}, z^{(2)}, ... , z^{(N)}\}$), as well as a list of permitted domains $\mathcal{Z} = (\mathcal{Z}_1, \mathcal{Z}_2, ... , \mathcal{Z}_{D_z})$ for each dimension of $z$ (e.g. a closed interval if $z_i$ is continuous, or a set of supported values if $z_i$ is discrete). Finally, assume we are given some distance metric $d(x, x')$ and threshold $\epsilon$.

The task consists of a sequence of $N_q$ questions. For each, we sample two representation values $z, z'$ iid from $p(Z)$, which have corresponding instance values $x = g(z)$ and $x' = g(z')$. In an interface, we visualize $x$ and $x'$ along with their distance $d(x,x')$ and its proximity to the threshold $\epsilon$. We also present manipulable controls that allow the user to modify each component of $z$ within $\mathcal{Z}$ to interactively change $x = g(z)$ and thus the distance $d(x,x')$. The user's goal is to manipulate the controls as efficiently as possible to bring $d(x,x') \leq \epsilon$, thereby reconstructing the target. Once this condition is achieved, we repeat the process for newly sampled $z$ and $z'$. If a user actively attempts to solve a question for a threshold amount of time $T$ to no avail, we permit them to move on.

During each question, we continuously record the values of $z$, the errors $d(x,x')$, the dimensions $i$ being changed, and the directions of modification (increasing or decreasing). These records let us precisely replay user actions and from them, derive a rich set of performance metrics, which we enumerate in Section~\ref{sec:fr-metrics}.
We hypothesize that, while the task will be possible to complete without understanding the model, users will perform it more reliably and efficiently when they intuitively understand what representation dimensions mean. We also hypothesize that the process of performing it will \textit{teach} users these intuitive meanings when they exist---that is, when the model is interpretable.

The above is a general definition of the interactive reconstruction task. To apply it to a specific problem, however, a number of implementation choices must be made:
\begin{itemize}
\item \textbf{Control of representations $z$}: There are many different ways of inputting values for representation dimensions (e.g., numeric fields vs. sliders for continuous dimensions, or radio buttons vs. dropdowns for discrete dimensions), and also many different ways of arranging and annotating them (e.g., allowing reordering, grouping, and labeling, which can be helpful for higher-dimensional $z$).
\item \textbf{Visualization of instances $x$:} Although some instance modalities have canonical visualizations (e.g., images), others can be visualized in many ways (e.g., patient medical records). Visualizations should make it easy to recognize the effects of changing $z$ and compare whether $x$ and $x'$ are becoming closer with respect to $d(\cdot,\cdot)$.
\item \textbf{Choice of distance metric $d(x,x')$, distance threshold $\epsilon$, and time threshold $T$: } These critical parameters are best chosen together. For our experiments, we relied on small studies for each dataset, seeking $d(\cdot,\cdot)$ and $\epsilon$ that captured when users subjectively felt they had manipulated $x$ to match $x'$ sufficiently closely, and setting $T$ to a round number on the order of twice the median duration.
\end{itemize}
We describe our dataset-specific choices in Section~\ref{sec:fr-impl}, but recommend these be re-tuned for new applications. 

\subsection{Theoretical Grounding for the Interactive Reconstruction Task} \label{sec:theoretical-grounding}
A number of sources in the HCI literature motivate and ground our interactive reconstruction task as a meaningful measure of understanding.  First, as recommended by Kulesza et al.~\cite{kulesza2013too}, our method attempts to present the ``whole truth'' and ``nothing but the truth:'' we visualize the entire model without any simplifying approximations. For generative models, visualizing the full model also means, to the greatest extent possible, visualizing dimensions jointly rather than separately (as is the case when using, e.g., latent traversals). We adopt this strategy in line with cognitive load theory as articulated by Sweller~\cite{sweller1994cognitive}, who argues that ``understanding'' emerges from explaining interactions between elements of a schema (``the basic unit of knowledge''), and that explaining interacting elements separately will make material harder to understand, rather than easier to understand, because of the split attention effect~\cite{chandler1992split}.

We also attempt to avoid split attention effects between the visualization and the proxy task. In many interpretability measurement methods, these two components are visually separate, e.g., in different portions of the screen~\cite{huysmans2011empirical,lage2018human,lage2019evaluation,poursabzi2018manipulating}. Such physical separation often makes it possible or even preferable to complete the task without engaging with the visualization, e.g., by guessing multiple-choice answers to complete the task more quickly. In recent HCI studies of interpretability tools, both Bu\c{c}inca et al.~\cite{buccinca2020proxy} and Kaur et al.~\cite{InterpretingInterpretability} note an inconsistency between outcomes in thinkaloud studies, where cognitive engagement with a visualization is forced, and practice, where the visualization can be readily ignored. By closely integrating the visualization and the task, we attempt to avoid this pitfall.

Finally, per Norman's articulation of the importance of feedback for building understanding in human-centered design~\cite{norman2013design}, our task is structured to provide immediate, interactive feedback. Just from looking at the screen, it is always readily apparent to users whether they have gotten the right answer, and whether they are getting closer or further away. In contrast, for most simulation-proxy interpretability measurement tasks~\cite{poursabzi2018manipulating,lage2019evaluation,huysmans2011empirical,slack2019assessing}, if users receive feedback at all, it is sporadic, e.g., about whether a multiple choice answer was correct.

\subsection{Baseline: the Single-Dimension Task}
\label{sec:sd-definition}

As a baseline to our interactive reconstruction task, we also consider a ``single-dimension'' task, inspired by existing interpretability measurement methods for discriminative models (Section~\ref{sec:hci-quant}).  Here, users are given an instance $x = g(z)$ and asked to guess the value of some hidden dimension $z_i$. To help them, users may view visualizations of other instances as dimension $i$ is varied, shown in a different section of the screen. For each model, users are asked $N_q$ questions about each dimension, and receive feedback after each question about whether their answer was correct. Details for our dataset-specific implementations are in Section~\ref{sec:sd-impl}.

The single-dimension task is deliberately designed to violate our theoretical motivations; specifically, (1) it does not present ``the whole truth'' of the model, (2) it visualizes dimensions individually rather than jointly, (3) it spatially separates the visualization from the task, and (4) it provides feedback sporadically (after each question) rather than interactively.

\subsection{Quantifying Quality of Interpretability Measurement Methods}
 \label{sec:meas-meas}

Measuring the accuracy of any measurement method generally requires testing with a precisely known quantity. 
In this paper, we \emph{assume ground-truth knowledge} that a particular model is more interpretable than another by working with synthetic datasets constructed for the express purpose of making this assumption reasonable, and supported by qualitative studies. We then test how well different interpretability measurement methods detect this assumed ground-truth difference. Where possible, we support our assumptions with quantitative measures of models' correspondence to ground-truth, e.g. disentanglement measures.

Our approach contrasts with the strategy of Bu\c{c}inca et al.~\cite{buccinca2020proxy}, who evaluate interpretability evaluation ``proxy tasks'' by how well they predict performance on downstream tasks. Although the ultimate motivation for interpretability research is to improve performance on downstream tasks ranging from better human+AI collaboration \cite{bansal2019beyond} to auditing for safety \cite{caruana2015intelligible}, 
performance on these tasks has no explicit correspondence to understanding. Additionally, in research settings, we often lack access to the true downstream tasks that motivate our work; thus our downstream validation task is \emph{also} a proxy. So in effect, we are not evaluating whether a proxy task measures interpretability, but whether one proxy predicts another.

\section{Implementation}  \label{sec:experimental-setup}

We now describe how we implement our approach for a variety of representation learning models and datasets.

\subsection{Datasets}
\begin{figure*}
    \centering
    \includegraphics[width=0.32\linewidth]{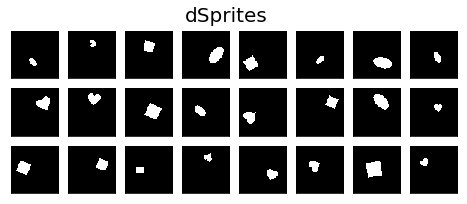}
    \includegraphics[width=0.32\linewidth]{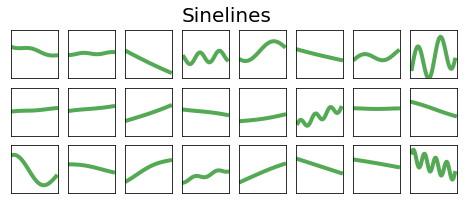}
    \includegraphics[width=0.32\linewidth]{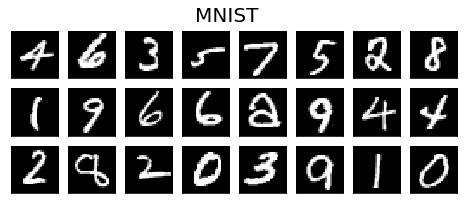}
    \caption{Examples from the dSprites, Sinelines, and MNIST datasets.}
    \label{fig:datasets}
    \Description[Sets of example images from the dSprites, Sinelines, and MNIST datasets.]{Three images, each showing 24 examples from the three different datasets used in the paper. The first, dSprites, consists of shapes with varying size and orientation. The second, Sinelines, consists of lines and waves. The third, MNIST, consists of handwritten digits.}
\end{figure*}

We considered three datasets, visualized in Figure \ref{fig:datasets}:
\begin{itemize}
\item \textbf{dSprites} \citep{dsprites17}, a 64$\times$64 binary image dataset with five ground-truth factors of generation $z$: shape, rotation, size, x-position, and y-position. We chose dSprites due to its popularity in the disentanglement literature (Section~\ref{sec:disentanglement}). 
\item \textbf{Sinelines}, a 64-dimensional timeseries dataset we developed for this study. Each instance $x = \langle x_1, x_2, ..., x_{64}\rangle$ is a mixture of a line and a sine wave, generated by mapping: 
  $x_t = z_1 t + z_2 + z_3 \sin(z_4 t + z_5),$
where $z_1 \sim \mathrm{Uniform}(-1,1)$ is slope, $z_2 \sim \mathcal{N}(0,1)$ is intercept, $z_3 \sim \mathrm{Exponential}(1)$ is amplitude, $z_4 \sim \mathrm{Exponential}(1)$ is frequency, and $z_5 \sim \mathrm{Uniform}(0, 2\pi)$ is phase. We make $z$ 5-dimensional for consistency with dSprites, but make $x$ a timeseries rather than an image to probe sensitivity to instance modality.
\item \textbf{MNIST}~\cite{lecun1998mnist}, a popular benchmark in the ML and interpretable representation learning literature consisting of images of handwritten digits from 0 to 9. Although the MNIST dataset lacks ground-truth representations, it does contain labels indicating depicted digits.
\end{itemize}

\subsection{Models}

On dSprites and Sinelines, we tested the following models:
\begin{itemize}
\item \textbf{``Ground-truth'' (GT)} generative models, constructed to have the same relationship between $x$ and $z$ as the (intentionally interpretable) process which generated the examples.
\item \textbf{Autoencoders (AE)} \cite{hinton2006reducing}, a baseline nonlinear representation learning method often considered to learn uninterpretable representations.
\item \textbf{Variational autoencoders (VAE)} \cite{kingma2013auto}, which are similar to autoencoders but learn distributions over $z$, with a prior on $z$ that can be interpreted as a regularizer. Because of this regularization effect, VAEs are widely reputed to learn more interpretable relationships between instances and representations than standard autoencoders.
\end{itemize}
Our primary goal was to validate that our task could distinguish entangled autoencoders from disentangled ground-truth models (both absolutely and relative to baselines).  

Architecturally, for dSprites, we used the same 7-layer convolutional neural networks (CNNs) as Burgess et al.~\cite{burgess2018understanding}, one of the original papers introducing dSprites (with the GT model just using the decoder). For Sinelines, we used 256$\times$256 fully connected networks with ReLU activations (except for the GT model, which was simple enough to implement in closed form). We tested all models at $D_z=5$ to match ground-truth.

On MNIST, which has no ground-truth model, our goal instead was to test a broader set of popular interpretable representation learning methods that have never been tested with user studies. We again tested on AEs and VAEs, but also included the following models from the disentanglement literature: 
\begin{itemize}
\item \textbf{$\beta$-TCVAEs (TC)}~\cite{chen2018isolating}, a modification of the VAE that is trained to learn representation dimensions $z_i$ that are statistically independent, generally considered near state of the art in the disentanglement literature.
\item \textbf{Semi-supervised $\beta$-TCVAEs (SS)}, a modified $\beta$-TCVAE we explicitly train to disentangle digit identity from style (in a discrete dimension). Though we lack full ground truth, we expect SS to be less entangled than TC.
\item \textbf{InfoGAN (IG)} \cite{chen2016infogan}, a generative adversarial \cite{goodfellow2014generative} disentanglement method that also learns to disentangle digit identity from digit style, but imperfectly and without supervision.
\end{itemize}
We tested all of these MNIST models (AE, VAE, TC, SS, and IG) at $D_z=5$ but additionally tested AE, TC, and SS at $D_z=10$ to probe sensitivity to representation dimensionality. Architecturally, for all models, we use the same CNN architecture given in the papers introducing InfoGAN~\cite{chen2016infogan} and $\beta$-TCVAE~\cite{chen2018isolating} (Table 4), changing only the size of the latent dimension $D_z$. Training details were chosen to match original specifications where possible; see Section~\ref{sec:arch-details} of the Appendix for additional details as well as model loss functions.

\begin{figure}
  \includegraphics[width=\linewidth]{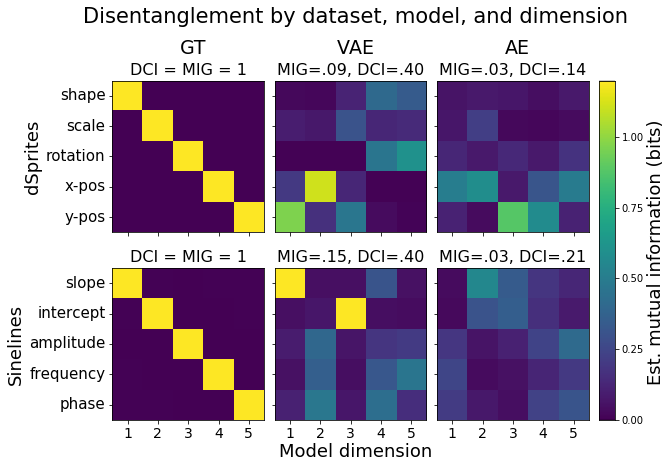}
  \caption{Disentanglement scores (in plot titles) and pairwise mutual information (in heatmaps, approximated using 2D histograms) between true generative factors and representation dimensions. By construction, ground-truth (GT) models are perfectly disentangled, while VAEs learn to partially concentrate information about certain ground-truth factors into individual representation dimensions. AEs exhibit less clear relationships and have the lowest disentanglement scores.}
  \label{fig:ground-truth-mutual-info}
  \Description[Disentanglement by dataset, model, and dimension.]{Heatmaps illustrating correspondence between learned dimensions and ground-truth dimensions for each of the three model types (GT, VAE, and AE) trained on the two synthetic datasets (dSprites and Sinelines). GT models correspond perfectly by construction. VAEs show some dimensions which mostly correspond but many that do not. AEs show little correspondence to ground-truth.}
\end{figure}

On dSprites and Sinelines, we quantified the extent to which our models matched ground truth with disentanglement measures. Specifically, we computed the DCI disentanglement score~\cite{eastwood2018framework} and the mutual information gap (MIG)~\cite{chen2018isolating}, which are commonly included in disentanglement papers. In addition to these overall scores, we computed pairwise mutual information to visualize dependence on a per-dimension basis. Figure~\ref{fig:ground-truth-mutual-info} shows each of these metrics for each dimension and dataset. On both datasets, GT models are perfectly disentangled and AE models are heavily entangled. VAEs are somewhere in the middle, partially disentangling horizontal and vertical position from shape, scale, and rotation on dSprites, and partially disentangling linear from sinusoidal factors on Sinelines. As mentioned previously in Section~\ref{sec:meas-meas}, we use these metrics to support our assumption that GT models are more interpretable than AEs.
On MNIST, we have no ground truth model, but we hypothesize that the semi-supervised (SS) model will be most interpretable due to disentanglement between its continuous dimensions and ground-truth digit identity.

\subsection{Interface} \label{sec:fr-impl}

\begin{figure*}[ht]
\centering
\fbox{\includegraphics[width=0.406\linewidth]{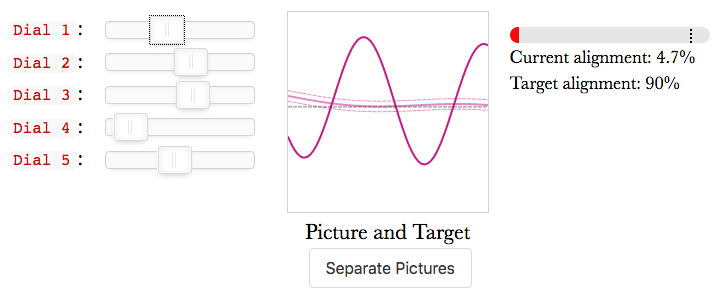}}
\fbox{\includegraphics[width=0.554\linewidth]{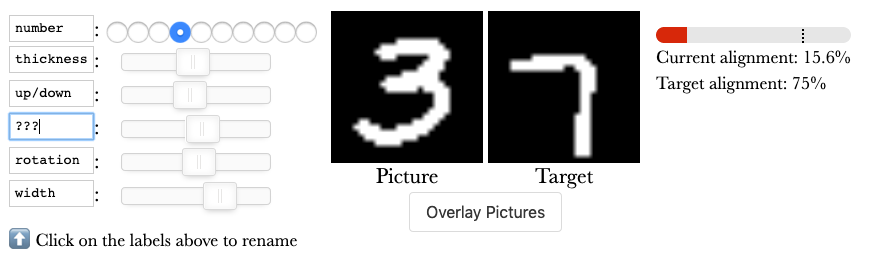}}
\caption{Screenshots of the interactive reconstruction task on Sinelines (left, with $x$ and $x'$ overlaid and dotted lines indicating the region of allowable alignment) and MNIST (right, with separated $x$ and annotations for $z$).} 
\label{fig:fr-screenshots}
\Description[Screenshots of interactive reconstruction.]{Two screenshot images showing the interactive reconstruction task. The left image is for Sinelines and shows overlaid output and target. The right image is for MNIST with output and target images separated.}
\end{figure*}

Figure~\ref{fig:fr-screenshots} shows examples of the user-facing interface for this method, implemented for two different datasets. The following interface elements correspond to the interactive reconstruction task parameters described in Section~\ref{sec:fr-definition}:

\textbf{Control of representations $z$}:
To specify closed-interval continuous dimensions of $z$, we use slider components, while to specify finite-support discrete dimensions, we use unlabeled radio buttons. We sampled initial $z$ and target $z'$ from $p(Z)$ set to the empirical distribution, or specifically a heldout testing split of the dataset. Slider ranges were determined by taking the empirical minimum and maximum values of each dimension $z_i$ over this heldout split. The one exception to this procedure was for the InfoGAN model, where sampling and limits were determined from the prior.

\textbf{Visualization of instances $x$}: For images, visualization was straightforward, while for Sinelines, we used line charts with dashed lines at $y=0$ and appropriate limits. In addition to visualizing $x$ and $x'$ side-by-side, we provided an option to overlay them with partial transparency, which we found was helpful in pilot experiments for fine-tuning. This defaulted to on for our synthetic experiments, but was defaulted to off for MNIST after pilot users expressed the side-by-side view was more helpful. For MNIST in particular, to facilitate remembering (and recording) user impressions of dimension meanings, we allowed users to input custom labels next to the corresponding controls in the interface.

\textbf{Distance and time thresholds}: For each dataset, we defined $d(x,x')$ as a Jaccard distance~\cite{levandowsky1971distance}, i.e., the fraction of disagreeing dimensions of $x$ and $x'$ to total active dimensions of $x$ and $x'$, with dataset-specific definitions of agreement and activity. Exact expressions are given in Section~\ref{sec:metrics-thresholds} of the Appendix. 
Although these choices worked well for black and white images and consistently-scaled timeseries, different metrics might be necessary for other data modalities; we discuss this further in Section~\ref{sec:discussion}. Because $d(x,x')$ was between 0 and 1, we visualized it as an agreement percentage rather than a distance, and chose $\epsilon=0.1$ (or a $90\%$ agreement threshold) for synthetic datasets and $\epsilon=0.25$ (or a $75\%$ agreement threshold) for MNIST.

The time threshold $T$ for skipping questions was set to 30 seconds for dSprites and Sinelines and 45 seconds for MNIST. We paused this hidden countdown whenever users were inactive for more than 3 seconds.

We implemented tasks as single-page, client-side web applications, with machine learning models running directly in users' web browsers after being converted to TensorFlow.js \cite{smilkov2019tensorflow}. Despite the fairly large size of certain models (e.g. 7-layer convolutional neural networks), only two users out of hundreds reported problems running them interactively in-browser. Links to the task for all models can be found at \url{http://hreps.s3.amazonaws.com/quiz/manifest.html}.

\subsection{Interactive Reconstruction Metrics}
\label{sec:fr-metrics}

We computed the following performance metrics from our records of users performing the interactive reconstruction task:

\begin{itemize}
\item \textbf{Completion rate:} For each model, we measure the fraction of questions the user solves, i.e. moves the sliders and/or chooses radio buttons to make $d(x,x') \leq \epsilon$, rather than skipping after $T$ seconds. 

\item \textbf{Response time:} For each model, we measure the average time it takes participants to complete each question. 

\item \textbf{Slide distance:} For each model, we measure the average total distance moved by sliders, in units of full slider widths, for each question. Discrete changes count as 1. 
\item \textbf{Error AUC:} We measure the ``area under the curve'' of mean squared difference between $x$ and $x'$ over the total time the user is attempting to solve the quiz.
\item \textbf{Self-reported difficulty:} After answering a ``stage'' of  questions about a model, users rate that difficulty from 1 to 7 using a Single Ease Question (SEQ)~\cite{sauro2009comparison}. 

\end{itemize}
We store snapshots of $x$ and $z$ every 100ms while the value is changing in a constant direction, or whenever the active direction or dimension changes. While this list is not an exhaustive enumeration of all possible metrics applicable to the task, others can be defined and computed post-hoc as we store a nearly complete record of user actions over time.

\subsection{Single-Dimension Task Parameters}
\label{sec:sd-impl}

In this section, we describe how we instantiated the single-dimension task 
for the synthetic datasets. This task has two primary implementation parameters: the prediction task and the visualization.

For the prediction task, we opted for a multiple-choice classification task where, for a particular example $x$ visualized in the same way as in the interactive reconstruction task, users decided whether $z_i$ is ``Low,'' ``Medium,'' or ``High.'' These regimes were respectively defined in terms of the 1st-5th, 48th-52nd, and 95th-99th percentiles of the marginal distribution of encoded $x$ in a held-out split of the original dataset. We sampled $z$ by first sampling from the empirical joint distribution (i.e. a heldout split of the dataset), then overriding $z_i$ to a value selected uniformly from one of these regimes. Users answered two questions for each dimension $i$, and received the correct answer as feedback.

\begin{figure*}[ht]
\centering
\fbox{\includegraphics[width=0.369\linewidth]{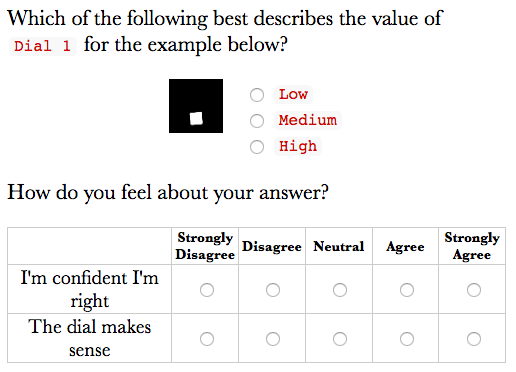}}
\fbox{\includegraphics[width=0.312625\linewidth]{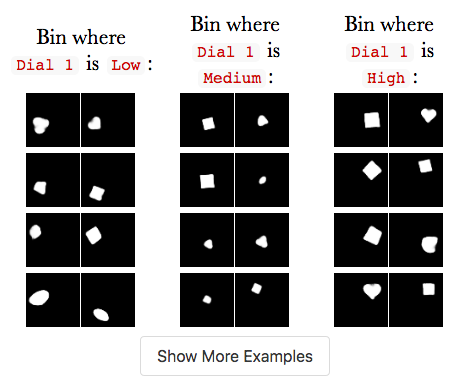}}
\fbox{\includegraphics[width=0.2545\linewidth]{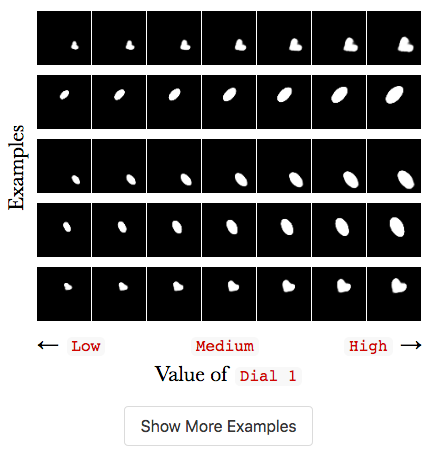}}
\caption{Screenshots of the dSprites single-dimension task (left), showing ground truth visualizations with exemplars (center) and traversals (right).
}
\label{fig:sd-screenshots}
\Description[Screenshots of the single-dimension task.]{Three screenshot images illustrating interactive reconstruction. The left image shows an example multiple-choice question with Likert scales for confidence. The middle image shows an exemplar visualization that could be used to answer the question. The right image shows a traversal visualization that could instead by used to answer the question.}
\end{figure*}

For the visualization, which is the sole task component dependent on the model, we tested two versions of the task, one using \textit{latent traversals} and the other using \textit{synthetic exemplars}, as described in Section~\ref{sec:disentanglement}.
For traversal visualizations, for 5 randomly sampled values of $z$, we plotted $x$ values corresponding to overriding $z_i$ to 7 linearly spaced values between ``Low'' and ``High'' as defined above. For exemplar visualizations, we showed  ``Low,'' ``Medium,'' and ``High'' bins with 8 examples each. Both of these visualizations force users to generalize from finite samples, which can lead to ambiguities if randomly sampled $z$ are not diverse. To mitigate this potential problem, we provided users with a ``Show More Examples'' button. Screenshots for the single-dimension task on the dSprites dataset are shown in Figure~\ref{fig:sd-screenshots}.

\subsection{Single-Dimension Metrics}
\label{sec:sd-metrics}

We recorded several performance metrics specific to the single-dimension task:
\begin{itemize}
\item \textbf{Correctness rate:} For each model, we measured the percentage of questions the participant answered correctly. 
\item \textbf{Self-reported confidence:} For each model, we measured whether users agreed with the statement ``I'm confident I'm right'' on a 5-point Likert scale. 
\item \textbf{Self-reported understanding:}
Similar Likert measurement, but for ``The dial makes sense.'' (We referred to dimensions as ``dials'' in the interface.) 
\end{itemize}
In addition to these task-specific metrics, we also recorded response time and self-reported difficulty with the SEQ, which were shared in common with interactive reconstruction.

\section{Study Methodology}

\begin{figure}
    \centering
    \includegraphics[width=\linewidth]{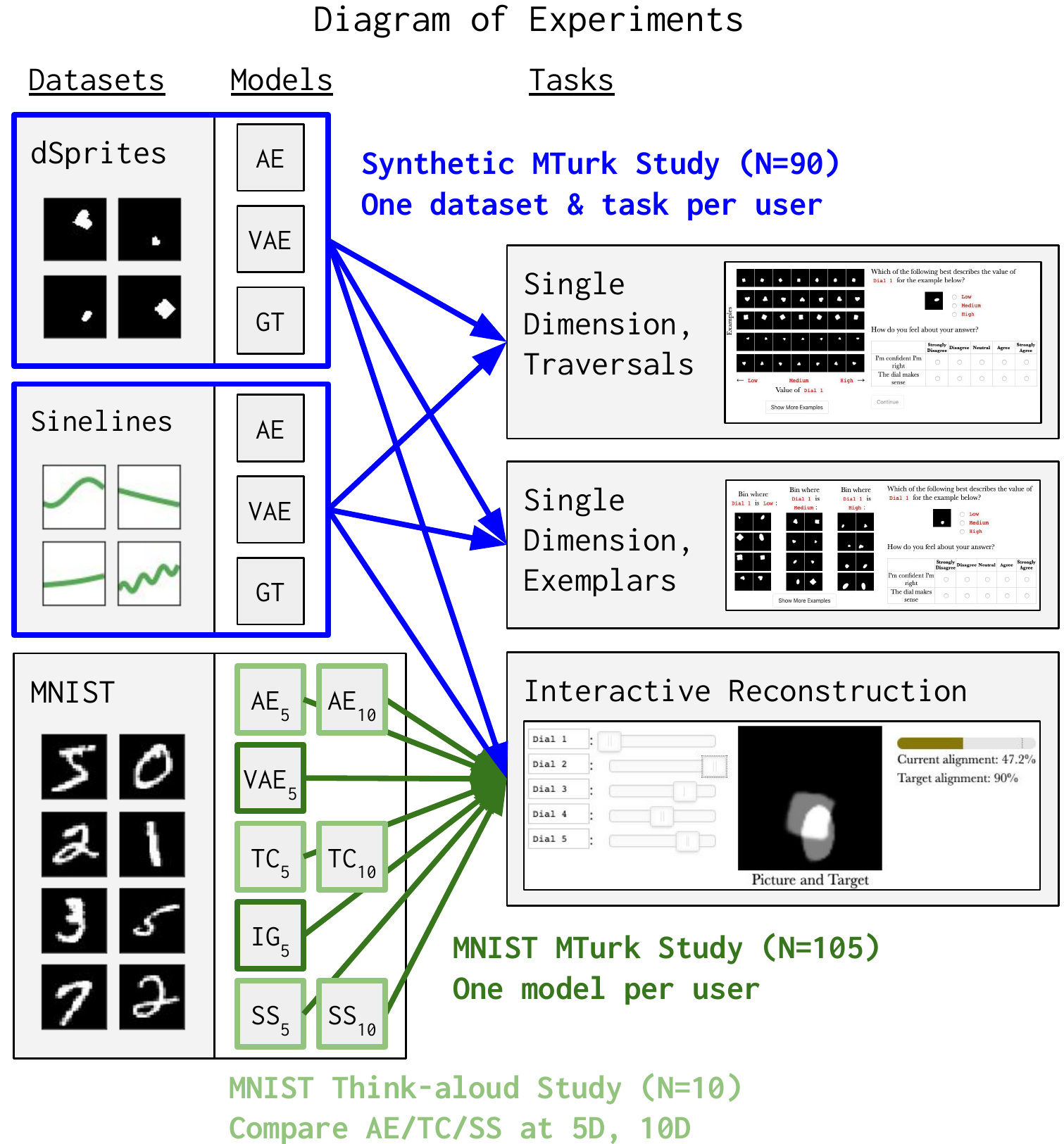}
    \caption{Diagram of experiments performed (except the small synthetic pilot). On synthetic datasets, experiments were designed to evaluate the best task (assuming the best model). On MNIST, experiments were designed to evaluate the best model (assuming the best task).}
    \label{fig:exp-design}
    \Description[Diagram of experiments.]{The image shows a diagram of which experiments in the study test which combinations of datasets, models, and tasks. Synthetic MTurk tests dSprites and Sinelines GT, VAE, and AE models against interactive reconstruction, single-dimension with exemplars, and single-dimension with traversals, with one dataset and task per user. MNIST MTurk tests seven models against interactive reconstruction, with one model per user. MNIST think-aloud tests three models per user against interactive reconstruction at either 5 or 10 dimensions.}
\end{figure}

With implementation decisions now specified, we describe the four studies we ran, two larger-scale studies on MTurk and two smaller-scale lab study sessions (see also Figure~\ref{fig:exp-design} for a graphical depiction). All experiments began with a consent form, a tutorial, and practice questions on an easy example. Experiments were deployed on the web and are available at \url{http://hreps.s3.amazonaws.com/quiz/manifest.html}. 
Code for our study is available at \url{https://github.com/dtak/interactive-reconstruction}.

\subsection{Experimental Design}

\subsubsection{Synthetic Think-aloud Pilot} We began by running a small pilot study with $N=3$ users (referred to as U1, U2, and U3) to test out each possible task on the dSprites and Sinelines synthetic datasets (interactive reconstruction, single-dimension with exemplars, and single-dimension with traversals), with each user completing two of the six task/dataset conditions with two of the six models (drawn randomly without replacement to ensure complete coverage). Participants were asked to think aloud and describe their strategies for solving each problem. Interviews were recorded and transcribed, and the feedback was used to make minor clarifying changes to the interface.

\subsubsection{Synthetic MTurk Study} We then ran a larger version of the synthetic dataset study on MTurk, where each participant ($N=15$ per dataset and task, and $N=90$ total) completed one of the six possible dataset/task conditions, but for all three models (AE, VAE, and GT, with the order randomized). To keep overall quiz length manageable, $N_q$ was set to 5 for interactive reconstruction and 10 for single-dimension tasks (that is, 2 questions per dimension). Differences were analyzed with repeated-measures ANOVA and paired t-tests.

\subsubsection{MNIST Think-aloud Study} We next ran a think-aloud study with $N=10$ participants on MNIST, testing out a wider variety of models but only for the interactive reconstruction task. Each participant (whom we refer to as P1-10) completed interactive reconstruction for three models (AE, TC, and SS, order drawn randomly without replacement), with odd and even-numbered participants working with 5D and 10D representations respectively. Model-specific stages ended after 7 questions had been completed or 10 minutes had elapsed. After each stage, in addition to the single ease question~\cite{sauro2009comparison}, users entered raw NASA-TLX scores \cite{hart1988development} and answered Likert scale questions about whether they felt they understood representation dimensions. Differences were analyzed with repeated-measures ANOVA and paired t-tests. During each stage, users were also encouraged to label dimensions if possible and continuously describe their impressions of the task, which we recorded.

\subsubsection{MNIST MTurk Study} Finally, we ran a large-scale MTurk study on MNIST. As with the synthetic dataset study, we had $N=15$ participants per condition (for $N=105$ participants total). However, instead of showing all models and varying the task, we limited the task to interactive reconstruction and varied the model, specifically showing each participant one of $\mathrm{AE}_5$, $\mathrm{VAE}_5$, $\mathrm{IG}_5$, $\mathrm{TC}_5$, $\mathrm{SS}_5$, $\mathrm{AE}_{10}$, or $\mathrm{SS}_{10}$. $N_q$ was increased from 5 to 7 for additional signal and to make the task sufficiently long. Differences were analyzed with one-way ANOVA and independent t-tests.

\subsection{Recruitment}

\subsubsection{Think-aloud Studies}
For our think-aloud studies, we recruited undergraduate and graduate students from computer science mailing lists at an academic institution, and compensated participants with Amazon gift cards (\$15/hour for synthetic pilot study sessions, which we increased to \$20/hour for MNIST to incentivize recruitment for a larger study). We recruited $N=3$ participants for the synthetic pilot and $N=10$ for MNIST.

On the synthetic datasets, two participants were male, one was female, and all were graduate students. On MNIST, two were male, eight were female, three were undergraduates, and seven were graduate students. All participants in both studies were aged 18-34.

\subsubsection{MTurk Studies}
For each MTurk experimental condition (6 for synthetic datasets and 7 for MNIST), we recruited $N=15$ participants on Amazon Mechanical Turk with unique worker IDs. This translates to a total of 90 participants on the synthetic dataset study and 105 participants for the MNIST-based study.

For single-dimension tasks, participants were excluded if they answered practice questions incorrectly.
For traversal visualizations, the retention rate was 71\%, while for exemplars, it was 91\%. For interactive reconstruction, we included all participants who successfully completed the study on synthetic datasets (as users could not proceed past the practice questions without answering them correctly), but added an additional inclusion criteria on MNIST requiring participants to have reconstructed at least one of seven instances successfully. We adopted this criteria in part because we noticed several participants did not realize they could use radio buttons (which were not included in the practice questions) on the IG and SS tasks (never changing them during any question), and in part to filter out users who gave up completely on extremely hard models (e.g. the 10D AE). Retention rates were highest (88\%) for the $\mathrm{SS}_{10}$ model and lowest (42\%) for the $\mathrm{AE}_{10}$, with an overall average of 67\%. In total, 156 participants were needed to get 105 retained participants on MNIST and 115 participants were needed to get 90 retained participants on synthetic tasks.

Compensation was set with the intention of ensuring that actively engaged MTurk users would earn at least \$15/hr, with time estimates based on pilot experiments and interactive reconstruction time limits. On the synthetic datasets, users were paid \$6 for both single-dimension and interactive reconstruction tasks, which took an average of 13 minutes (\$27/hr). The interactive reconstruction experiments took a slightly longer average of 18 minutes (\$20/hr), with 95\% of participants finishing under 34 minutes (\$11/hr).
For the MNIST dataset, users were paid \$3.75 for completing tasks, which took an average of 15 minutes (\$15/hr), with 80\% of participants finishing under 20 minutes (\$12/hr) and 95\% finishing under 38 minutes (\$6/hr).

Demographic information was recorded in a post-quiz questionnaire. In the synthetic study, participants were generally young (58\% between 18-34),  North American (68\%, with most others from Asia or South America), college-educated (79\% had Bachelor's degrees or above) and male (64\%), with gender recorded per guideline G-4 of \cite{scheuerman2019hci}.
Demographics were similar on MNIST (49\% aged 18-34, 59\% North American, 71\% male, and 81\% with Bachelor's or above).

\section{Results}
Since the think-aloud and MTurk study participants worked with the same datasets and completed similar or identical tasks, we will first present results from all the synthetic dataset studies and then move on to results from the MNIST dataset studies. Adding up ANOVAs and t-tests across all our experiments and pairs of models, we ran 435 statistical tests, giving us a Bonferroni-corrected threshold of $\signifThresh$ for an initial $\alpha$ of 0.05. Means, standard deviations, and pairwise p-values for all metrics across all experiments are given in Section~\ref{sec:full-results} of the Appendix.

\subsection{Synthetic Study Results}

\subsubsection{MTurk, Interactive Reconstruction} 
\begin{figure*}
    \centering
    \includegraphics[width=0.33\textwidth]{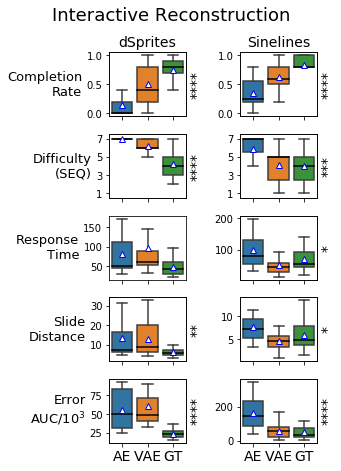}
    \includegraphics[width=0.33\textwidth]{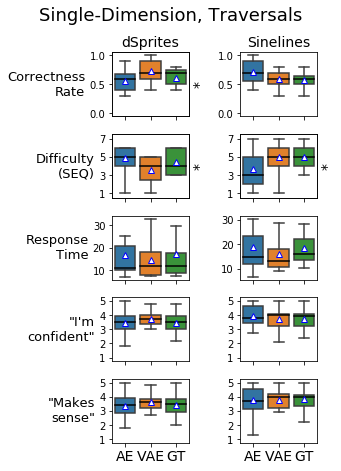}
    \includegraphics[width=0.33\textwidth]{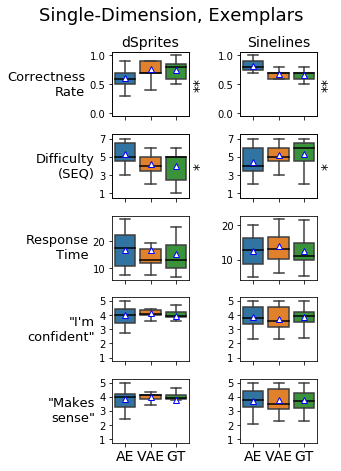}
    \caption{Boxplots of MTurk dependent variables across the 15 participants for each synthetic data task condition. Blue triangles indicate means, black lines indicate medians, and stars indicate p-values for differences between means (****=p<0.0001, ***=p<0.001, **=p<0.01, *=p<0.05). Expected differences between AE and GT models emerge clearly from interactive reconstruction results, but not single-dimension results. The same results are repeated as tables in Table~\ref{tbl:synth-all} of the Appendix.}
    \label{fig:boxplots}
    \Description[Boxplots showing MTurk task metrics on dSprites and Sinelines.]{Three boxplots show MTurk task metrics on dSprites and Sinelines for AE vs. VAE vs. GT models for interactive reconstruction (left), single-dimension with traversals (middle), and single-dimension with exemplars (right). Interactive reconstruction metrics are much better at differentiating GT models from the other two.}
\end{figure*}

As shown in Figure~\ref{fig:boxplots}, 
\textbf{the interactive reconstruction task clearly differentiated ground truth (GT) models from AEs.} Completion rates were significantly different on both dSprites ($F_{2,28}=32.3, p{<}0.0001$) and Sinelines ($F_{2,28}{=}23.4, p{<}0.0001$) in the directions we hypothesized. GT model completion rates on both dSprites ($.75{\pm}.26$) and Sinelines ($.83{\pm}.19$) were significantly higher than AE completion rates ($.15{\pm}.19$, $t_{14}{=}8.5$, $p{<}0.0001$ on dSprites and $.34{\pm}{.26}$, $t_{14}{=}6.4$, $p{<}0.0001$ on Sinelines). VAE models, though only marginally significantly different, were in the middle ($.51{\pm}.33$ on dSprites and $.63{\pm}.28$ on Sinelines). These results closely mirror the disentanglement metrics in Figure~\ref{fig:ground-truth-mutual-info}.

We also saw significant differences in self-reported difficulty ($F_{2,28}=32.3, p{<}0.0001$ on dSprites, $F_{2,28}{=}11.7, p{=}0.0002$ on Sinelines) and error AUC ($F_{2,28}{=}31.6$, $p{<}0.0001$ on dSprites, $F_{2,28}{=}14.1$, $p{<}0.0001$ on Sinelines). Between AE and GT models, these differences were large and least marginally significant. VAEs were no longer exactly in the middle, however. Instead, they more closely matched the AE model on dSprites and the GT model on Sinelines---e.g. for difficulty, average scores for the AE, VAE, and GT were 6.9, 6.4, and 4.2 on dSprites vs. 5.9, 4.1, and 4.0 on Sinelines.

\subsubsection{Think-aloud, Interactive Reconstruction}
\textbf{Interactive reconstruction task users felt they understood GT models, did not understand AEs, and partially understood VAEs} in a way that matched the disentanglement scores in Figure \ref{fig:ground-truth-mutual-info}.
On dSprites, U1 assigned meanings to GT model dimensions almost immediately, noting on the first question that \textit{``[dial 1] is changing its shape, [...] dial 2 is changing the size, [...] and then 3 seems like it's changing the rotation, and 4, probably x-position. And then 5 is y-direction I guess?''} For the AE, however, U1 \textit{``d[id]n't understand what the dials [we]re doing.''} On Sinelines, U2 found the GT model \textit{``much easier to figure out''} than the AE, where it was \textit{``hard to tease out exactly what each [dimension] is doing.''} The difficulty of understanding the AE model caused them to switch strategies from \textit{''match[ing] [dimensions] sequentially''} (which they could do with the GT model) and instead \textit{``look[ing] at the current alignment and then mak[ing] sure that number is increasing''} by \textit{''tweak[ing] the dials and see[ing] what happens.''}

Users partially understood VAEs. On the Sinelines VAE, U2 reported they could \textit{``quickly figure out''} there \textit{``was one dial that [controlled] slope''} and another that \textit{``did [vertical] translation''}, but that \textit{``squiggliness''} was \textit{''quite hard.''} These reports match the dependence plots in Figure \ref{fig:ground-truth-mutual-info}, which show that the Sinelines VAE disentangled linear slope and intercept from sine wave parameters.
On the dSprites VAE, U1 felt \textit{``dials 1, 2 and 3 are more predictable''} after determining \textit{``dial 3 is changing the size''} while \textit{``dials 1 and 2 are moving the object in the diagonal way.''} These comments again matched the dependence plots in Figure~\ref{fig:ground-truth-mutual-info}, which suggest position and size are controlled by VAE dimensions 1-3, while shape and rotation are controlled by dimensions 4-5. However, U1 did find the \textit{``diagonal''} relationship (which was also nonlinear) confusing, which forced them to randomly experiment more within each group of dimensions. In contrast, for U2, linear slope and intercept were disentangled both from sine wave dimensions and each other. This difference may explain why, relative to the GT, VAEs had higher MTurk difficulty ratings and error AUCs on dSprites than Sinelines (despite having similar DCI scores).

\subsubsection{MTurk, Single-Dimension}
\begin{figure}
    \centering
    \includegraphics[width=\linewidth]{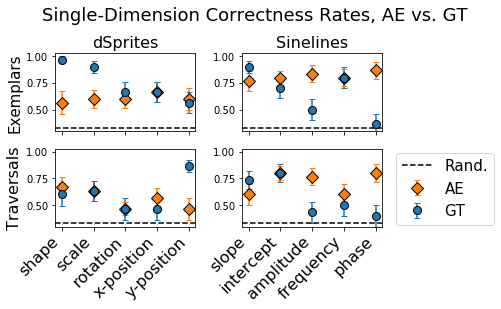}
    \caption{AE and GT correctness rates by representation dimension for MTurk single-dimension tasks. Dots show means with standard errors, x-axis shows GT dimensions (AE order is arbitrary). Compared to AEs, some GT dimensions had much lower or higher correctness rates than others, rising to near 100\% for attributes like shape, scale, and slope and falling to near 33\% (the rate of random guessing) for others, especially periodic attributes like rotation and phase. These differences suggest that questions about certain conceptually simple dimensions were pathologically difficult to answer from static visualizations.}
    \label{fig:dimdiffdiff}
    \Description[Single-dimension correctness rates for AE vs. GT]{The image shows the correctness rates associated with different individual dimensions for GT vs. AE models, plotted as dots with standard error errorbars. Some GT dimensions had very high correctness rates while others had very low correctness rates. AE correctness rates were less variable across dimensions.}
\end{figure}

\textbf{The single-dimension task did not clearly differentiate models.} In the MTurk results, no differences in any metrics were significant at 
$\alpha = \signifThresh$, though correctness rates for the exemplar-based version of the task were closest ($F_{2,28}{=}5.8$, $p{=}0.008$ for dSprites and $F_{2,28}{=}6.6$, $p{=}0.005$ for Sinelines, though the most significant relationships were in the wrong direction). Response time and self-reported confidence/understanding were almost identical across models, datasets, and visualizations. 
On dSprites, we did observe that GT models had higher correctness rates ($.75{\pm}.14$) and lower difficulty ratings ($4.0{\pm}1.6$) than AEs, which had $.61{\pm}.18$, $p{=}.01$ for correctness and $5.3{\pm}1.3$, $p{=}.03$ for difficulty (on exemplars). 
However, on Sinelines, AEs actually emerged with the \emph{highest} average correctness rates in the exemplar-based MTurk study ($\mathrm{AE}{=}.81{\pm}.15$ vs. $\mathrm{GT}{=}.65{\pm}.11$, $p{=}0.003$), as well as the lowest difficulty ($\mathrm{AE}{=}4.5{\pm}1.6$ vs. $\mathrm{GT}{=}5.3{\pm}1.5$, $p{=}0.003$). 
 Our qualitative results below (as well as the dimension-by-dimension correctness rate breakdown in Figure~\ref{fig:dimdiffdiff}) suggest this is not because users understood the AE, but because of helpful AE pathologies and unhelpful GT symmetries. 

\subsubsection{Think-aloud, Single-Dimension}
\textbf{Users performed the task without trying to understand the model.} Although U1 initially tried to formulate \textit{``hypotheses''} about the meanings of dimensions, they found these \textit{``irrelevant for [them] to make decisions.''} U2 described their strategy as \textit{``match[ing] directly''} without trying to \textit{``tease apart the different dimensions,''} and P3 called it \textit{``blind trust''} in \textit{``visual matching.''}  

Visualization pathologies also biased the results in directions unrelated to understanding. On Sinelines, we noticed a lack of diversity in ``High'' and ``Low'' samples from AEs, which sometimes made the matching problem trivially easy for all participants. For GT models, we also noticed questions for particular were pathologically hard due to symmetries in the generative process. One example is that for the rotation feature on dSprites and the phase feature on Sinelines, ``Low'' values near 0 and ``High'' values near $2\pi$ were nearly indistinguishable. For GT models on Sinelines, participants also found it hard to detect changes in amplitude at low frequencies, as these could be alternately explained by slope, and harder still to detect changes in frequency at low amplitudes, as near-linear examples were effectively unaltered. 

\subsubsection{Exemplars vs. Traversals.}
\label{sec:exem-vs-trav}
Though not central to our narrative, we found some evidence that exemplars may have been more effective than traversals for single-dimension performance. U3, who completed versions with both visualizations, reported that the task was easier with exemplars because seeing instances clustered into bins \textit{``visually help[ed them] understand the three categories,''} whereas with the traversal visualization, it was necessary to \textit{``mentally create''} those categories by imagining spatial \textit{``divisions.''} U3 also hypothesized it was easier for them to detect patterns on the right-hand side of traversal visualizations due to their experience \textit{``reading from left to right,''} which they were concerned might introduce \textit{``a bias''} in their answers. We see such evidence of spatial bias in Figure~\ref{fig:dimdiffdiff}, whose bottom-left plot shows that MTurk users struggled to answer traversal questions about dSprites x-position, where the movement of the shape was in the same direction as the movement of the images, but answered questions about y-position almost perfectly, where the movement of the shape was orthogonal to the movement of images. Meanwhile, accuracy on these questions for exemplars (top-left plot) was intermediate in both cases, showing less negative bias for x-position though also less positive bias for y-position. Overall, these results suggest that visualization details matter for performance, but also that it may be better to utilize time rather than space when visualizing changes in spatial features, which is effectively what occurs with interactivity.

\subsection{MNIST Study Results}
\label{sec:mnist-results}

\subsubsection{MTurk, Interactive Reconstruction} 
\begin{figure*}
\includegraphics[width=\textwidth]{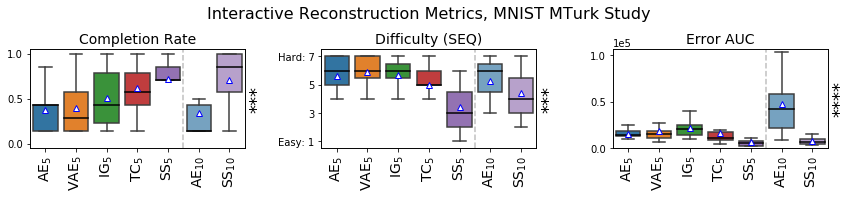}
\caption{Boxplots across the 15 MTurk participants for each MNIST model, for dependent variables that were significant in the synthetic tasks. Blue triangles indicate means, black lines indicate medians, and stars indicate significance as in Figure~\ref{fig:boxplots}. As with the qualitative studies, semi-supervised (SS) models performed best by each measure, while standard autoencoders (AE) performed near the worst, especially at higher representation dimensions (right). Pairwise comparisons can be found in Figure~\ref{fig:fr-pairwise} of the Appendix.}
\label{fig:mnist-mturk}
\Description[Boxplots for MNIST MTurk results.]{Boxplots showing completion rate, subjective difficulty, and error AUC for the seven different MNIST models tested on MTurk. The semi-supervised models perform the best across all categories while the AE models perform worst.}
\end{figure*}

As shown in Figure~\ref{fig:mnist-mturk},
\textbf{Interactive reconstruction metrics clearly distinguished models, especially AE vs. SS.} In general, we found significant differences between models ($F_{28}{=}4.7, p{=}0.0003$ for completion rate, $F_{28}{=}4.9, p{=}0.0002$ for difficulty, and $F_{28}{=}11.4, p{<}0.0001$ for error AUC), with the sharpest pairwise contrasts being significant or nearly significant differences between AE and SS models. For example, at 5D for the AE vs. SS, completion rates were $0.38\pm 0.23$ vs. $0.72\pm 0.24$ ($t_{28}{=}{-}3.9$, $p{=}0.0006$), subjective difficulty was $5.6\pm1.6$ vs. $3.4\pm1.5$ ($t_{28}{=}3.7$, $p{=}0.001$), and error AUC (given in units of 1000s for conciseness) was $15.7\pm4.3$ vs. $6.3\pm 4.7$ ($t_{28}{=}5.5$, $p{<}0.0001$). The next-best performing model was the $\beta$-TCVAE (TC), with an average completion rate of $0.61\pm0.27$, subjective difficulty of $5.0\pm1.5$, and error AUC of $16.0\pm14.0$. Although the average error AUC was higher for the TC model than the AE, this was largely due to a small number of extreme outliers (note the high variance). Comparing medians, we have 10.5 for the TC model and 14.5 for the AE model, which matches the overall trend of AE < TC < SS. Results for the IG and especially the VAE were generally slightly worse and closer to AE performance.

\textbf{Performance degraded when increasing dimensionality, but not as badly for methods thought to be interpretable.} When we increased $D_z$ from 5 to 10, average completion rate fell ($0.38{\to}0.35$ for the AE, $0.73{\to}0.71$ for the SS), error AUC rose ($15.7{\to}47.8$ for the AE, $6.3{\to}7.6$ for the SS), and subjective difficulty generally rose ($3.4{\to}4.4$ for the SS, though it fell slightly from $5.6{\to}5.3$ for the AE), though none of these results were near significance except the AE's increase in error AUC ($t{=}{-}3.7$, $p{=}0.002$). Examining medians, however, we find that median completion rate fell dramatically for the AE ($3/7{\to}1/7,$ the minimum possible value for retention), while for the SS it actually \emph{rose} ($5/7{\to} 6/7$), suggesting that degradation from increasing dimensionality was more drastic for the AE than the SS.

\subsubsection{Think-aloud, Interactive Reconstruction}
\begin{figure*}
\includegraphics[width=\textwidth]{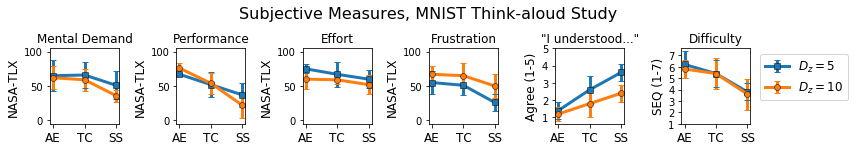}
\caption{Subjective measures for the MNIST think-aloud study (means $\pm$ standard deviations). The first four measures are NASA-TLX scores \cite{hart1988development}, the fifth is users' subjective agreement (from 1-5) with the statement ``I understood what many of the dimensions meant,'' and the last is a single ease question (SEQ) \cite{sauro2009comparison} assessment of difficulty from 1-7. Users rated SS models easiest across all measures, though for higher $D_z$, subjective understandability was lower, frustration was higher, and perceived performance gaps were greater. An alternate presentation can be found in Figure~\ref{fig:fr-pairwise} of the Appendix.}
\label{fig:mnist-subjective-measures}
\Description[Subjective measures for the MNIST think-aloud study.]{Six line plots showing (1) NASA-TLX mental demand, (2) performance, (3) effort, (4) frustration, (5) Likert-scale subjective understanding, and (6) single ease question difficutly for AE vs. TC. vs SS MNIST models at 5 vs 10 dimensions during the thinkaloud study. SS performs best, AE performs worst.}
\end{figure*}

\textbf{Multiple sources of evidence suggest users understood the SS well and the AE poorly}.
The first source is subjective measures, which are plotted in Figure~\ref{fig:mnist-subjective-measures}. Though not always significant at $N=5$, they suggest not just that the difficulty and cognitive load of the task varied across models (with AE generally rated hardest and SS easiest), but also that dimension understandability varied in the same way, e.g., with SS rated most understandable at 5D with $3.6{\pm}{0.5}$ and AE rated least understandable with $1.4{\pm}0.5$ ($t_{4}{=}11.0$, $p{=}0.0004$). When increasing $D_z$ to 10, frustration grew and subjective understanding of SS models fell (to $2.4{\pm}0.5$, $t_{4}{=}6$, $p{=}0.004$), but relative differences remained fairly consistent.

The second source of evidence is agreement between user-entered dimension labels, shown for $D_z=5$ in Figure \ref{fig:mnist-labeling-5d}. On average, only 20\% of users were able to assign labels to any AE dimensions, but 100\% were able to do so for at least one SS or TC dimension, with all dimensions at $D_z=5$ and many at $D_z=10$ having at least 80\% coverage. Additionally, for many SS dimensions, these independently-assigned labels agreed closely. TC model labels were less consistent, suggesting more entanglement with the digit.

\begin{table*}[t]
\begin{tabular}{|c|c|l|}
\hline
Model &     Dimension &                                                       User-Assigned Labels \\\hline\hline
 \multirow{5}{*}{$\mathrm{AE}_5$} &  Continuous 1 &   \texttt{"791"} \\\cline{2-3}
  &  Continuous 2 &   \texttt{"thickness"} \\\cline{2-3}
  &  Continuous 3 &   \texttt{"rotate"} \\\cline{2-3}
  &  Continuous 4 &   \texttt{"stretch"} \\\cline{2-3}
  &  Continuous 5 &  ---\\\hline\hline
 \multirow{5}{*}{$\mathrm{TC}_5$} &  Continuous 1 &   \texttt{"thickness", "0-1", "closedness", "2-0"} \\\cline{2-3}
  &  Continuous 2 &   \texttt{"shrinking horizontally", "paint white", "0-1"} \\\cline{2-3}
  &  Continuous 3 &   \texttt{\textbf{"diagonal"}, \textbf{"rotate"}, "4-5-3"} \\\cline{2-3}
  &  Continuous 4 &   \texttt{"add bottom left", \textbf{"3-5-6"}, \textbf{"3-5-blob"}} \\\cline{2-3}
  &  Continuous 5 &   \texttt{"ccw", \textbf{"6-7"}, \textbf{"7-6?"}} \\\hline\hline
 \multirow{6}{*}{$\mathrm{SS}_5$} &  Discrete 1   &   \texttt{\textbf{"number"}, \textbf{"Digit"}, \textbf{"Class"}, \textbf{"digit"}} \\\cline{2-3}
  &  Continuous 1 &   \texttt{"up", "Widthish", "paint white", "rotate cw, change focus"} \\\cline{2-3}
  &  Continuous 2 &  \texttt{\textbf{"ccw"}, \textbf{"LR Skew"}, \textbf{"Rotation"}, \textbf{"rotate"}, \textbf{"skew rotate cw/ccw"}} \\\cline{2-3}
  &  Continuous 3 &   \texttt{"curve", \textbf{"TB Skew"}, \textbf{"up down bias"}, \textbf{"focus up/dwn"}} \\\cline{2-3}
  &  Continuous 4 &   \texttt{"thickness", \textbf{"Wide"}, \textbf{"width"}, \textbf{"horiz thickness somewhat"}} \\\cline{2-3}
  &  Continuous 5 &   \texttt{"left", \textbf{"Thick"}, \textbf{"paint black"}, \textbf{"line thickness/focus but white"}} \\\hline
\end{tabular}

\caption{All labels assigned to $D_z=5$ models by participants in the MNIST thinkaloud study. Bold text shows labels experimenters identified as consistent between participants. The semi-supervised (SS) model had both the most labels and the most consistent labels, while the autoencoder (AE) had the fewest. Labels for $D_z=10$ are given in Table~\ref{fig:mnist-labeling-10d} in the Appendix.}
\label{fig:mnist-labeling-5d}
\end{table*}

The third source of evidence is users' verbal descriptions. None of the 10 participants made any comments indicating that they found the AE comprehensible, with P3 commenting that ``\textit{the dials didn't have any discernible meaning}'', and P8 concluding that ``\textit{the meanings of the dials aren't helpful in this one for solving the problem most efficiently}.''
In contrast, 9/10 participants felt that the SS model was ``\textit{the easiest to understand and label}'' (P3), with 100\% agreeing it was more comprehensible than the AE. This was in large part ``\textit{because it [...] let you choose the number}'' (P4, with all participants commenting on this in some manner), though 8/10 participants also expressed that the continuous dimensions were at least partially understandable. For example, P1 noted that they ``\textit{understood what most of the dimensions were doing, especially rotating}.''
However, unlike with the synthetic GT models, no participants claimed complete understanding; for example, P4 felt ``\textit{there were some dials that were easy to tell, but there were others that were more obscure}.''

The TC model consistently fell between the others, with 8/10 commenting that it was more comprehensible than the AE, and 9/10 commenting that it was less comprehensible than the SS. Subjective descriptions varied; compared to the AE, P9 felt able to ``\textit{understand and write down and remember}'' what TC dials meant, despite being ``\textit{worried}'' that ``\textit{the framework [they] built might not be accurate}.'' P7 felt they could ``\textit{understand physically}'' what certain dimensions were ``\textit{trying to do}'' but had trouble expressing it verbally; they described it as being like a ``\textit{matching game}'' where ``\textit{you know what you mean but there’s not an agreed upon word for it}.''

\textbf{Differences in understanding led to differences in strategy.} When users understood dimensions, they used that understanding to perform the task more efficiently. On the SS model, P5 ``\textit{used the dials that [they] understood first}'' and ``\textit{had an idea of which way to move}'' them, while P8 felt it was ``\textit{pretty easy to know which [SS] dial to pick}.'' P7 was able ``\textit{to be intuitive}'' when they ``\textit{could figure out what the dials meant}.'' P1 even felt a degree of mastery, saying that for the SS model, they ``\textit{had it down to a more exact science}.''

In contrast, when users did not understand dimensions, they gravitated towards an inefficient but less cognitively taxing strategy similar to gradient ascent.
P3 described this strategy as trying to ``\textit{watch one slider and look at the alignment number of see if it reaches a local max},'' with some participants ``\textit{proceeding through dials one at a time}'' (P10) and others selecting them in ``\textit{a very random order}'' (P8). Users generally did not generally want to resort to this strategy; for the AE, P9 started off by putting in significant effort ``\textit{trying to see if [they] could find any understanding of what the dials meant}'' but ``\textit{it didn't work},'' and with reluctance, their ``\textit{strategy changed}'' to ``\textit{finicking with the dials}.'' However, P8 felt that for complicated models, the gradient ascent strategy was actually ``\textit{less mentally demanding}'' because ``\textit{not even trying to figure out}'' dimension meanings meant the task was sufficiently mindless that they could ``\textit{do something else at the same time},'' such as ``\textit{hold a conversation}.''
Users felt this was inefficient, with P2 commenting that
``\textit{this is definitely not the fastest strategy},'' but when ``\textit{there wasn't any easy way to define anything [...] to do it systematically feels really annoying}'' (P8).
These comments may help to explain why the differences in subjective measures of effort were less significant than differences in frustration, understanding, and performance.

\newcommand{\viewIcon}{\textcolor{gray}{\ArrowBoldUpRight}}

\begin{table*}[h!]
  \centering
  \begin{tabular}{|c|c|r|c|c|c|c|c|c|}\hline
  \multicolumn{3}{|c|}{Model Information} &  \multicolumn{2}{c|}{Disentanglement}
  &
   \multicolumn{4}{c|}{Interactive Reconstruction}

   \\\hline
    Dataset & Model & MSE & DCI~\cite{eastwood2018framework}& MIG~\cite{chen2018isolating} & Completion Rate & Difficulty & Error AUC/$10^3$ & Link \\\hline\hline
    \multirow{3}{*}{dSprites} & $\mathrm{AE}_5$  & \textbf{6.5} & 0.14 & 0.03 & 0.15±0.19 & 6.93±0.25 & 55.8±25.6 & \href{http://hreps.s3.amazonaws.com/quiz/index.html?viz=sliders&dataset=dsprites&skip_instructions=1&anonymized=1&models=conv_ae}{\viewIcon} \\\cline{2-9}
     & $\mathrm{VAE}_5$  & 8.1 & 0.40 & 0.09 & 0.51±0.33 & 6.20±0.91 & 60.1±26.6 & \href{http://hreps.s3.amazonaws.com/quiz/index.html?viz=sliders&dataset=dsprites&skip_instructions=1&anonymized=1&models=conv_vae}{\viewIcon} \\\cline{2-9}
     & $\mathrm{GT}_5$ & 8.9 & \textbf{1.00} & \textbf{1.00} &\textbf{0.75±0.26} & \textbf{4.20±1.60} & \textbf{23.6±5.9} & \href{http://hreps.s3.amazonaws.com/quiz/index.html?viz=sliders&dataset=dsprites&skip_instructions=1&anonymized=1&models=conv_supervised}{\viewIcon}\\\hline \hline

   \multirow{3}{*}{Sinelines} & $\mathrm{AE}_5$            & 0.6 & 0.21 & 0.03 & 0.34±0.26 & 5.87±1.67 & 163.5±90.0  & \href{http://hreps.s3.amazonaws.com/quiz/index.html?viz=sliders&dataset=timeseries&skip_instructions=1&anonymized=1&models=fc_ae}{\viewIcon}\\\cline{2-9}
     & $\mathrm{VAE}_5$           & 3.3 & 0.40 & 0.15 & 0.63±0.28 & 4.07±1.84 & 59.6±47.8 & \href{http://hreps.s3.amazonaws.com/quiz/index.html?viz=sliders&dataset=timeseries&skip_instructions=1&anonymized=1&models=fc_vae}{\viewIcon}\\\cline{2-9}
    & $\mathrm{GT}_5$  & \textbf{0.0} & \textbf{1.00} & \textbf{1.00} & \textbf{0.83±0.19} &\textbf{4.00±1.71} & \textbf{52.0±48.9} & \href{http://hreps.s3.amazonaws.com/quiz/index.html?viz=sliders&dataset=timeseries&skip_instructions=1&anonymized=1&models=supervised}{\viewIcon}\\\hline\hline

    \multirow{5}{*}{MNIST} & $\mathrm{AE}_5$ & \textbf{15.5} & --- & --- & 0.38±0.23 & 5.60±1.62 & 15.7±4.3 & \href{http://hreps.s3.amazonaws.com/quiz/index.html?viz=sliders&dataset=mnist&skip_instructions=1&anonymized=1&models=conv_ae_5}{\viewIcon} \\\cline{2-9}
     & $\mathrm{VAE}_5$ & 16.2 & --- & --- & 0.40±0.31 & 5.87±1.15 & 18.6±13.7 & \href{http://hreps.s3.amazonaws.com/quiz/index.html?viz=sliders&dataset=mnist&skip_instructions=1&anonymized=1&models=conv_vae_5}{\viewIcon}  \\\cline{2-9}
     & $\mathrm{IG}_5$ & ---$\,\,\,$ & --- & --- & 0.51±0.31 & 5.67±1.30 & 21.3±8.6 & \href{http://hreps.s3.amazonaws.com/quiz/index.html?viz=sliders&dataset=mnist&skip_instructions=1&anonymized=1&models=infogan-6}{\viewIcon} \\\cline{2-9}
     & $\mathrm{TC}_5$ & 25.4 & --- & --- & 0.62±0.26 & 5.00±1.41 & 16.0±13.1 & \href{http://hreps.s3.amazonaws.com/quiz/index.html?viz=sliders&dataset=mnist&skip_instructions=1&anonymized=1&models=conv_tcvae_9.0_K5}{\viewIcon}  \\\cline{2-9}
     & $\mathrm{SS}_5$ & 20.8 & --- & --- & \textbf{0.73±0.24} & \textbf{3.40±1.54} & \textbf{6.3±4.7} & \href{http://hreps.s3.amazonaws.com/quiz/index.html?viz=sliders&dataset=mnist&skip_instructions=1&anonymized=1&models=conv_semisupervised_vae_K-5_tc-9.0}{\viewIcon}   \\\hline\hline
    \multirow{3}{*}{MNIST} & $\mathrm{AE}_{10}$ & \textbf{7.2} & --- & --- & 0.35±0.29 & $\,$5.27±1.73 & 47.8±34.0 & \href{http://hreps.s3.amazonaws.com/quiz/index.html?viz=sliders&dataset=mnist&skip_instructions=1&anonymized=1&models=conv_ae_10}{\viewIcon}  \\\cline{2-9}
     & $\mathrm{TC}_{10}$ & 24.9 & --- & --- & --- & --- & ---   & \href{http://hreps.s3.amazonaws.com/quiz/index.html?viz=sliders&dataset=mnist&skip_instructions=1&anonymized=1&models=conv_tcvae_9.0_K10}{\viewIcon}\\\cline{2-9}
     & $\mathrm{SS}_{10}$ & 20.7 & --- & --- & \textbf{0.71±0.29} & \textbf{4.40±1.40} & \textbf{7.6±3.3}  & \href{http://hreps.s3.amazonaws.com/quiz/index.html?viz=sliders&dataset=mnist&skip_instructions=1&anonymized=1&models=conv_semisupervised_vae_K-10_tc-9.0}{\viewIcon} \\\hline
  \end{tabular}
  \caption{MTurk interactive reconstruction metric means and standard deviations along with general metrics for each of the models used in all experiments. Mean squared error (MSE) measures autoencoders' errors reconstructing inputs not seen during training. DCI and MIG are measures of disentanglement applicable to synthetic datasets. Bold entries indicate least error / greatest interpretability or disentanglement in each group. Links go directly to tasks for each model (skipping instructions).}
  \label{tbl:model-metrics}
\end{table*}

\begin{figure*}
\includegraphics[width=0.33\textwidth]{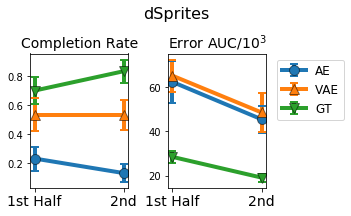}
\includegraphics[width=0.33\textwidth]{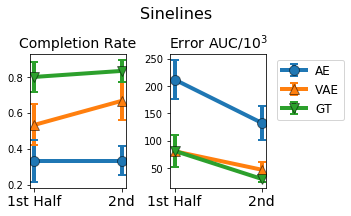}
\includegraphics[width=0.33\textwidth]{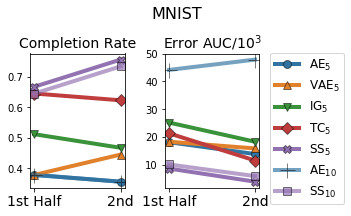}
\caption{
Changes in average MTurk interactive reconstruction metrics between the first and last $(N_q-1)/2$ questions (error-bars show standard error and are omitted on MNIST for readability). GT/SS models show the most consistent improvements over questions (suggesting conscious effort / learning) while AEs sometimes show degradation (suggesting users give up).}
\label{fig:ch-ch-changes}
\Description[Changes in reconstruction metrics from the first half to the last half of questions.]{Three images showing how metrics changed from the first half to the second half of MTurk interactive reconstruction questions on dSprites (left), Sinelines (middle), and MNIST (right). GT and SS models always did better in the second half, while AE models often did worse.}
\end{figure*}

\section{Discussion}\label{sec:discussion}

As shown in Figure~\ref{fig:boxplots} and Table~\ref{tbl:model-metrics},
\textbf{interactive reconstruction metrics differentiated entangled and disentangled models, both absolutely and relative to baselines.} On both synthetic datasets, these metrics (specifically completion rate, difficulty rating, and error AUC) clearly distinguished the entangled AE from the disentangled GT model. Single-dimension task metrics, on the other hand, showed a much less clear relationship with disentanglement, sometimes predicting that the highly entangled autoencoder was more interpretable than ground truth. On MNIST, interactive reconstruction metrics continued to meaningfully differentiate models in a manner consistent with hypotheses from the disentangled representations literature (e.g. $\beta$-TCVAEs especially semi-superivsed $\beta$-TCVAEs were more interpretable than AEs and standard VAEs).

\textbf{Interactive reconstruction metrics measured understanding.} 
This is a major claim, but our results suggest that performing the task helped users understand models, and that understanding models helped users perform the task.  First, we know that after performing interactive reconstruction, users felt they understood the ground-truth interpretable models much more than unregularized autoencoder models.
On MNIST, where there was no GT model, users held similar feelings towards the SS vs. the AE. Although one could attribute these feelings to an illusion of explanatory depth~\cite{rozenblit2002misunderstood}, that would not explain why user feelings were consistent for the same pairs of models or why users assigned similar labels to dimensions of models they felt they understood (Table~\ref{fig:mnist-labeling-5d}). It also would not explain evidence of learning across trials, e.g. why performance increased most consistently for interpretable models (Figure~\ref{fig:ch-ch-changes}). Overall, the evidence suggests that users achieved meaningfully different levels of model understanding by performing the interactive reconstruction (and not the single-dimension) task.

Second, we found that when users felt they understood the meanings of dimensions, they could modify them in a single pass through the list, knowing in advance in which direction and how much to change them. This allowed them to solve problems efficiently, leading directly to high completion rates and low error AUC. When users did not understand dimensions, we found that they tended to adopt less efficient gradient ascent or random experimentation procedures with numerous loops through the full set of dimensions (or a partial set, in the intermediate-understanding case).

A competing hypothesis could be that, rather than gaining understanding, users \emph{always} used gradient ascent or experimented randomly, but just happened to more easily stumble upon solutions with the models we assumed were ``interpretable'' than the models we assumed were ``uninterpretable.'' In certain cases, this effect may be partially operative. For example, on MNIST, AEs consistently have the lowest reconstruction error (shown in Table~\ref{tbl:model-metrics}), implying they can reconstruct a wider variety of images and therefore have a larger ``search space.'' Meanwhile, TCs consistently have the highest reconstruction error ($25.4$ vs. $15.5$ at $D_z=5$) and therefore the smallest search space, so randomly experimenting users ought to happen upon solutions more quickly.

However, this hypothesis is at odds with our qualitative observations, and also conflicts with our quantitative results in many cases. For example, on MNIST, the SS model has lower reconstruction error (20.8 at $D_z=5$) and thus a larger search space than the TC model, but it performs much better. On Sinelines, the GT model has precisely zero reconstruction error but still performs best. Although having a smaller search space may be helpful (and arguably less expressive models may often be easier to understand), we posit that task performance depends more strongly on the understandability of a search space, rather than its size.

\textbf{Generalizability.}
It is worth emphasizing that interactive reconstruction differentiated models in qualitatively similar ways across multiple datasets (dSprites, Sinelines, and MNIST) and user groups (workers on Amazon Mechanical Turk and students in Computer Science).
This consistency suggests that interactive reconstruction can be useful as a research tool for comparing interpretable representation learning methods in varied contexts.
Some innovation (beyond choosing appropriate parameters per Section~\ref{sec:fr-definition}) may still be required to adapt the method to non-visual or discontinuous data modalities (e.g. audio, text, or medical records) or to specific user groups in application-grounded contexts (per Doshi-Velez and Kim~\cite{doshi2017towards}).
However, while it is desirable for interpretable representation learning algorithms to generalize to all contexts, interpretability \emph{measurement} tools can afford to be a little more domain-specific, as long as they can still identify the algorithms that output the most interpretable models across contexts.

\textbf{Limitations.}  Interactive reconstruction has limitations not shared by the single-dimension task which our experiments do not fully explore. First, these tasks require interactive visualization, which may be technically challenging for practitioners to implement, though the increasing usability of efficient web-based machine learning frameworks \cite{smilkov2019tensorflow} helps. Second, interacting with all dimensions simultaneously may be overwhelming for models with many tens or hundreds of representation dimensions. Potential workarounds include allowing users to annotate, as with MNIST, and/or group dimensions, or defining tasks over subgroups rather than the full set. However, we note that 100D representations may be inherently uninterpretable due to limits on working memory~\cite{sweller1994cognitive}---unless the models are structured so that only sparse subsets of dimensions need to be considered simultaneously.

Our task also requires setting several parameter choices sensibly.  For example, despite our initial pilots on MNIST, 8/10 think-aloud study participants made at least one comment that our distance metric, the intersection-over-union alignment percentage, did not match their intuitive notion of perceptual similarity, with P3 commenting that sometimes the images ``\textit{look similar but the alignment [percentage] doesn't reflect that}.'' Although it was uncommon for users to reach the threshold alignment $\epsilon$ without feeling that the images were similar, they felt frustrated that changes in $d(x,x')$ seemed unrelated to progress early on. Although we feel confident our method will generalize to many datasets and data types, our experience suggests practitioners will need to take care when selecting metrics $d(x,x')$ and thresholds $\epsilon$. Future work could explore choosing example-specific thresholds or, on visual data, using metrics explicitly designed to model perceptual similarity~\cite{wang2004image,zhang2018unreasonable,czolbe2020loss} (if they can be evaluated efficiently in-browser).

Finally, interactive reconstruction is specific to generative models, e.g. autoencoders and GANs. Single-dimension tasks, however, can be made to support the other main category of representations, embeddings, via feature visualization~\cite{olah2017feature}. Evaluating embedding interpretability is an area for future work. Cavallo et al.~\cite{cavallo2018visual}, Smilkov et al.~\cite{smilkov2016embedding}, and Arendt et al.~\cite{arendt2020parallel} may provide a foundation.

\section{Conclusion}

Developing reliable methods of evaluating interpretability is important for progress in interpretable ML. In this study, we introduced an interactive reconstruction task for evaluating the interpretability of generative models, which have largely gone unstudied in the growing literature on human factors in ML.
We validated our method by verifying it was effective at identifying ground-truth differences in model interpretability---both absolutely and relative to baselines---and that differences in objective performance metrics corresponded to meaningful differences in subjective understanding, which was measured in multiple independent ways.
We then applied it to a wide range of representation learning methods from the disentanglement literature, and found evidence that methods which have been shown to improve disentanglement on synthetic data, e.g., \cite{chen2018isolating}, also improve interpretability on real data. To our awareness, ours is the first study providing such evidence.

\section*{Acknowledgments}

We thank Krzysztof Gajos, Zana Bu\c{c}inca, Zilin Ma, Giulia Zerbini, Benjamin Levy, and the Harvard DtAK group for helpful discussions and insights. ASR and FDV acknowledge support from NIH 1R56MH115187 and the Sloan Fellowship.

\bibliographystyle{ACM-Reference-Format}
\bibliography{bibliography}


\begin{thebibliography}{73}


\ifx \showCODEN    \undefined \def \showCODEN     #1{\unskip}     \fi
\ifx \showDOI      \undefined \def \showDOI       #1{#1}\fi
\ifx \showISBNx    \undefined \def \showISBNx     #1{\unskip}     \fi
\ifx \showISBNxiii \undefined \def \showISBNxiii  #1{\unskip}     \fi
\ifx \showISSN     \undefined \def \showISSN      #1{\unskip}     \fi
\ifx \showLCCN     \undefined \def \showLCCN      #1{\unskip}     \fi
\ifx \shownote     \undefined \def \shownote      #1{#1}          \fi
\ifx \showarticletitle \undefined \def \showarticletitle #1{#1}   \fi
\ifx \showURL      \undefined \def \showURL       {\relax}        \fi
\providecommand\bibfield[2]{#2}
\providecommand\bibinfo[2]{#2}
\providecommand\natexlab[1]{#1}
\providecommand\showeprint[2][]{arXiv:#2}

\bibitem[\protect\citeauthoryear{Abdul, von~der Weth, Kankanhalli, and
  Lim}{Abdul et~al\mbox{.}}{2020}]%
        {abdul2020cogam}
\bibfield{author}{\bibinfo{person}{Ashraf Abdul}, \bibinfo{person}{Christian
  von~der Weth}, \bibinfo{person}{Mohan Kankanhalli}, {and}
  \bibinfo{person}{Brian~Y Lim}.} \bibinfo{year}{2020}\natexlab{}.
\newblock \showarticletitle{COGAM: Measuring and Moderating Cognitive Load in
  Machine Learning Model Explanations}. In
  \bibinfo{booktitle}{\emph{Proceedings of the 2020 CHI Conference on Human
  Factors in Computing Systems}}. \bibinfo{pages}{1--14}.
\newblock


\bibitem[\protect\citeauthoryear{Ackley, Hinton, and Sejnowski}{Ackley
  et~al\mbox{.}}{1985}]%
        {ackley1985learning}
\bibfield{author}{\bibinfo{person}{David~H Ackley}, \bibinfo{person}{Geoffrey~E
  Hinton}, {and} \bibinfo{person}{Terrence~J Sejnowski}.}
  \bibinfo{year}{1985}\natexlab{}.
\newblock \showarticletitle{A learning algorithm for Boltzmann machines}.
\newblock \bibinfo{journal}{\emph{Cognitive science}} \bibinfo{volume}{9},
  \bibinfo{number}{1} (\bibinfo{year}{1985}), \bibinfo{pages}{147--169}.
\newblock


\bibitem[\protect\citeauthoryear{Allahyari and Lavesson}{Allahyari and
  Lavesson}{2011}]%
        {allahyari2011user}
\bibfield{author}{\bibinfo{person}{Hiva Allahyari} {and}
  \bibinfo{person}{Niklas Lavesson}.} \bibinfo{year}{2011}\natexlab{}.
\newblock \showarticletitle{User-oriented assessment of classification model
  understandability}. In \bibinfo{booktitle}{\emph{11th scandinavian conference
  on Artificial intelligence}}. IOS Press.
\newblock


\bibitem[\protect\citeauthoryear{Alvarez-Melis and Jaakkola}{Alvarez-Melis and
  Jaakkola}{2018}]%
        {alvarez2018towards}
\bibfield{author}{\bibinfo{person}{David Alvarez-Melis} {and}
  \bibinfo{person}{Tommi~S Jaakkola}.} \bibinfo{year}{2018}\natexlab{}.
\newblock \showarticletitle{Towards robust interpretability with
  self-explaining neural networks}.
\newblock \bibinfo{journal}{\emph{arXiv preprint arXiv:1806.07538}}
  (\bibinfo{year}{2018}).
\newblock


\bibitem[\protect\citeauthoryear{Arendt, Nur, Huang, Fair, and Dou}{Arendt
  et~al\mbox{.}}{2020}]%
        {arendt2020parallel}
\bibfield{author}{\bibinfo{person}{Dustin~L Arendt}, \bibinfo{person}{Nasheen
  Nur}, \bibinfo{person}{Zhuanyi Huang}, \bibinfo{person}{Gabriel Fair}, {and}
  \bibinfo{person}{Wenwen Dou}.} \bibinfo{year}{2020}\natexlab{}.
\newblock \showarticletitle{Parallel embeddings: a visualization technique for
  contrasting learned representations}. In
  \bibinfo{booktitle}{\emph{Proceedings of the 25th International Conference on
  Intelligent User Interfaces}}. \bibinfo{pages}{259--274}.
\newblock


\bibitem[\protect\citeauthoryear{Bansal, Nushi, Kamar, Lasecki, Weld, and
  Horvitz}{Bansal et~al\mbox{.}}{2019}]%
        {bansal2019beyond}
\bibfield{author}{\bibinfo{person}{Gagan Bansal}, \bibinfo{person}{Besmira
  Nushi}, \bibinfo{person}{Ece Kamar}, \bibinfo{person}{Walter~S Lasecki},
  \bibinfo{person}{Daniel~S Weld}, {and} \bibinfo{person}{Eric Horvitz}.}
  \bibinfo{year}{2019}\natexlab{}.
\newblock \showarticletitle{Beyond accuracy: The role of mental models in
  human-AI team performance}. In \bibinfo{booktitle}{\emph{Proceedings of the
  AAAI Conference on Human Computation and Crowdsourcing}}.
  \bibinfo{pages}{2--11}.
\newblock


\bibitem[\protect\citeauthoryear{Bau, Zhou, Khosla, Oliva, and Torralba}{Bau
  et~al\mbox{.}}{2017}]%
        {bau2017network}
\bibfield{author}{\bibinfo{person}{David Bau}, \bibinfo{person}{Bolei Zhou},
  \bibinfo{person}{Aditya Khosla}, \bibinfo{person}{Aude Oliva}, {and}
  \bibinfo{person}{Antonio Torralba}.} \bibinfo{year}{2017}\natexlab{}.
\newblock \showarticletitle{Network dissection: Quantifying interpretability of
  deep visual representations}. In \bibinfo{booktitle}{\emph{Proceedings of the
  IEEE conference on computer vision and pattern recognition}}.
  \bibinfo{pages}{6541--6549}.
\newblock


\bibitem[\protect\citeauthoryear{Bau, Zhu, Strobelt, Zhou, Tenenbaum, Freeman,
  and Torralba}{Bau et~al\mbox{.}}{2019}]%
        {bau2019visualizing}
\bibfield{author}{\bibinfo{person}{David Bau}, \bibinfo{person}{Jun-Yan Zhu},
  \bibinfo{person}{Hendrik Strobelt}, \bibinfo{person}{Bolei Zhou},
  \bibinfo{person}{Joshua~B Tenenbaum}, \bibinfo{person}{William~T Freeman},
  {and} \bibinfo{person}{Antonio Torralba}.} \bibinfo{year}{2019}\natexlab{}.
\newblock \showarticletitle{Visualizing and Understanding GANs}.
\newblock  (\bibinfo{year}{2019}).
\newblock


\bibitem[\protect\citeauthoryear{Beals}{Beals}{1984}]%
        {beals1984bray}
\bibfield{author}{\bibinfo{person}{Edward~W Beals}.}
  \bibinfo{year}{1984}\natexlab{}.
\newblock \showarticletitle{Bray-Curtis ordination: an effective strategy for
  analysis of multivariate ecological data}.
\newblock In \bibinfo{booktitle}{\emph{Advances in ecological research}}.
  Vol.~\bibinfo{volume}{14}. \bibinfo{publisher}{Elsevier},
  \bibinfo{pages}{1--55}.
\newblock


\bibitem[\protect\citeauthoryear{Bengio, Courville, and Vincent}{Bengio
  et~al\mbox{.}}{2013}]%
        {bengio2013representation}
\bibfield{author}{\bibinfo{person}{Yoshua Bengio}, \bibinfo{person}{Aaron
  Courville}, {and} \bibinfo{person}{Pascal Vincent}.}
  \bibinfo{year}{2013}\natexlab{}.
\newblock \showarticletitle{Representation learning: A review and new
  perspectives}.
\newblock \bibinfo{journal}{\emph{IEEE transactions on pattern analysis and
  machine intelligence}} \bibinfo{volume}{35}, \bibinfo{number}{8}
  (\bibinfo{year}{2013}), \bibinfo{pages}{1798--1828}.
\newblock


\bibitem[\protect\citeauthoryear{Blei, Ng, and Jordan}{Blei
  et~al\mbox{.}}{2003}]%
        {blei2003latent}
\bibfield{author}{\bibinfo{person}{David~M Blei}, \bibinfo{person}{Andrew~Y
  Ng}, {and} \bibinfo{person}{Michael~I Jordan}.}
  \bibinfo{year}{2003}\natexlab{}.
\newblock \showarticletitle{Latent dirichlet allocation}.
\newblock \bibinfo{journal}{\emph{Journal of machine Learning research}}
  \bibinfo{volume}{3}, \bibinfo{number}{Jan} (\bibinfo{year}{2003}),
  \bibinfo{pages}{993--1022}.
\newblock


\bibitem[\protect\citeauthoryear{Breiman}{Breiman}{2001}]%
        {breiman2001random}
\bibfield{author}{\bibinfo{person}{Leo Breiman}.}
  \bibinfo{year}{2001}\natexlab{}.
\newblock \showarticletitle{Random forests}.
\newblock \bibinfo{journal}{\emph{Machine learning}} \bibinfo{volume}{45},
  \bibinfo{number}{1} (\bibinfo{year}{2001}), \bibinfo{pages}{5--32}.
\newblock


\bibitem[\protect\citeauthoryear{Bu{\c{c}}inca, Lin, Gajos, and
  Glassman}{Bu{\c{c}}inca et~al\mbox{.}}{2020}]%
        {buccinca2020proxy}
\bibfield{author}{\bibinfo{person}{Zana Bu{\c{c}}inca}, \bibinfo{person}{Phoebe
  Lin}, \bibinfo{person}{Krzysztof~Z Gajos}, {and} \bibinfo{person}{Elena~L
  Glassman}.} \bibinfo{year}{2020}\natexlab{}.
\newblock \showarticletitle{Proxy tasks and subjective measures can be
  misleading in evaluating explainable AI systems}. In
  \bibinfo{booktitle}{\emph{Proceedings of the 25th International Conference on
  Intelligent User Interfaces}}. \bibinfo{pages}{454--464}.
\newblock


\bibitem[\protect\citeauthoryear{Burgess, Higgins, Pal, Matthey, Watters,
  Desjardins, and Lerchner}{Burgess et~al\mbox{.}}{2018}]%
        {burgess2018understanding}
\bibfield{author}{\bibinfo{person}{Christopher~P Burgess},
  \bibinfo{person}{Irina Higgins}, \bibinfo{person}{Arka Pal},
  \bibinfo{person}{Loic Matthey}, \bibinfo{person}{Nick Watters},
  \bibinfo{person}{Guillaume Desjardins}, {and} \bibinfo{person}{Alexander
  Lerchner}.} \bibinfo{year}{2018}\natexlab{}.
\newblock \showarticletitle{Understanding disentangling in beta-VAE}.
\newblock \bibinfo{journal}{\emph{arXiv preprint arXiv:1804.03599}}
  (\bibinfo{year}{2018}).
\newblock


\bibitem[\protect\citeauthoryear{Cai, Reif, Hegde, Hipp, Kim, Smilkov,
  Wattenberg, Viegas, Corrado, Stumpe, et~al\mbox{.}}{Cai
  et~al\mbox{.}}{2019}]%
        {cai2019human}
\bibfield{author}{\bibinfo{person}{Carrie~J Cai}, \bibinfo{person}{Emily Reif},
  \bibinfo{person}{Narayan Hegde}, \bibinfo{person}{Jason Hipp},
  \bibinfo{person}{Been Kim}, \bibinfo{person}{Daniel Smilkov},
  \bibinfo{person}{Martin Wattenberg}, \bibinfo{person}{Fernanda Viegas},
  \bibinfo{person}{Greg~S Corrado}, \bibinfo{person}{Martin~C Stumpe},
  {et~al\mbox{.}}} \bibinfo{year}{2019}\natexlab{}.
\newblock \showarticletitle{Human-centered tools for coping with imperfect
  algorithms during medical decision-making}. In
  \bibinfo{booktitle}{\emph{Proceedings of the 2019 CHI Conference on Human
  Factors in Computing Systems}}. \bibinfo{pages}{1--14}.
\newblock


\bibitem[\protect\citeauthoryear{Caruana, Lou, Gehrke, Koch, Sturm, and
  Elhadad}{Caruana et~al\mbox{.}}{2015}]%
        {caruana2015intelligible}
\bibfield{author}{\bibinfo{person}{Rich Caruana}, \bibinfo{person}{Yin Lou},
  \bibinfo{person}{Johannes Gehrke}, \bibinfo{person}{Paul Koch},
  \bibinfo{person}{Marc Sturm}, {and} \bibinfo{person}{Noemie Elhadad}.}
  \bibinfo{year}{2015}\natexlab{}.
\newblock \showarticletitle{Intelligible models for healthcare: Predicting
  pneumonia risk and hospital 30-day readmission}. In
  \bibinfo{booktitle}{\emph{Proceedings of the 21th ACM SIGKDD international
  conference on knowledge discovery and data mining}}.
  \bibinfo{pages}{1721--1730}.
\newblock


\bibitem[\protect\citeauthoryear{Cavallo and Demiralp}{Cavallo and
  Demiralp}{2018}]%
        {cavallo2018visual}
\bibfield{author}{\bibinfo{person}{Marco Cavallo} {and}
  \bibinfo{person}{{\c{C}}a{\u{g}}atay Demiralp}.}
  \bibinfo{year}{2018}\natexlab{}.
\newblock \showarticletitle{A visual interaction framework for dimensionality
  reduction based data exploration}. In \bibinfo{booktitle}{\emph{Proceedings
  of the 2018 CHI Conference on Human Factors in Computing Systems}}.
  \bibinfo{pages}{1--13}.
\newblock


\bibitem[\protect\citeauthoryear{Chandler and Sweller}{Chandler and
  Sweller}{1992}]%
        {chandler1992split}
\bibfield{author}{\bibinfo{person}{Paul Chandler} {and} \bibinfo{person}{John
  Sweller}.} \bibinfo{year}{1992}\natexlab{}.
\newblock \showarticletitle{The split-attention effect as a factor in the
  design of instruction}.
\newblock \bibinfo{journal}{\emph{British Journal of Educational Psychology}}
  \bibinfo{volume}{62}, \bibinfo{number}{2} (\bibinfo{year}{1992}),
  \bibinfo{pages}{233--246}.
\newblock


\bibitem[\protect\citeauthoryear{Chen, Li, Grosse, and Duvenaud}{Chen
  et~al\mbox{.}}{2018}]%
        {chen2018isolating}
\bibfield{author}{\bibinfo{person}{Ricky~TQ Chen}, \bibinfo{person}{Xuechen
  Li}, \bibinfo{person}{Roger~B Grosse}, {and} \bibinfo{person}{David~K
  Duvenaud}.} \bibinfo{year}{2018}\natexlab{}.
\newblock \showarticletitle{Isolating sources of disentanglement in variational
  autoencoders}. In \bibinfo{booktitle}{\emph{Advances in neural information
  processing systems}}. \bibinfo{pages}{2610--2620}.
\newblock


\bibitem[\protect\citeauthoryear{Chen, Duan, Houthooft, Schulman, Sutskever,
  and Abbeel}{Chen et~al\mbox{.}}{2016}]%
        {chen2016infogan}
\bibfield{author}{\bibinfo{person}{Xi Chen}, \bibinfo{person}{Yan Duan},
  \bibinfo{person}{Rein Houthooft}, \bibinfo{person}{John Schulman},
  \bibinfo{person}{Ilya Sutskever}, {and} \bibinfo{person}{Pieter Abbeel}.}
  \bibinfo{year}{2016}\natexlab{}.
\newblock \showarticletitle{Infogan: Interpretable representation learning by
  information maximizing generative adversarial nets}. In
  \bibinfo{booktitle}{\emph{Advances in neural information processing
  systems}}. \bibinfo{pages}{2172--2180}.
\newblock


\bibitem[\protect\citeauthoryear{Czolbe, Krause, Cox, and Igel}{Czolbe
  et~al\mbox{.}}{2020}]%
        {czolbe2020loss}
\bibfield{author}{\bibinfo{person}{Steffen Czolbe}, \bibinfo{person}{Oswin
  Krause}, \bibinfo{person}{Ingemar Cox}, {and} \bibinfo{person}{Christian
  Igel}.} \bibinfo{year}{2020}\natexlab{}.
\newblock \showarticletitle{A Loss Function for Generative Neural Networks
  Based on Watson’s Perceptual Model}.
\newblock \bibinfo{journal}{\emph{Advances in Neural Information Processing
  Systems}}  \bibinfo{volume}{33} (\bibinfo{year}{2020}).
\newblock


\bibitem[\protect\citeauthoryear{Desjardins, Courville, and Bengio}{Desjardins
  et~al\mbox{.}}{2012}]%
        {desjardins2012disentangling}
\bibfield{author}{\bibinfo{person}{Guillaume Desjardins},
  \bibinfo{person}{Aaron Courville}, {and} \bibinfo{person}{Yoshua Bengio}.}
  \bibinfo{year}{2012}\natexlab{}.
\newblock \showarticletitle{Disentangling factors of variation via generative
  entangling}.
\newblock \bibinfo{journal}{\emph{arXiv preprint arXiv:1210.5474}}
  (\bibinfo{year}{2012}).
\newblock


\bibitem[\protect\citeauthoryear{Doshi-Velez and Kim}{Doshi-Velez and
  Kim}{2017}]%
        {doshi2017towards}
\bibfield{author}{\bibinfo{person}{Finale Doshi-Velez} {and}
  \bibinfo{person}{Been Kim}.} \bibinfo{year}{2017}\natexlab{}.
\newblock \showarticletitle{Towards a rigorous science of interpretable machine
  learning}.
\newblock \bibinfo{journal}{\emph{arXiv preprint arXiv:1702.08608}}
  (\bibinfo{year}{2017}).
\newblock


\bibitem[\protect\citeauthoryear{Eastwood and Williams}{Eastwood and
  Williams}{2018}]%
        {eastwood2018framework}
\bibfield{author}{\bibinfo{person}{Cian Eastwood} {and}
  \bibinfo{person}{Christopher~KI Williams}.} \bibinfo{year}{2018}\natexlab{}.
\newblock \showarticletitle{A framework for the quantitative evaluation of
  disentangled representations}. In \bibinfo{booktitle}{\emph{International
  Conference on Learning Representations}}.
\newblock


\bibitem[\protect\citeauthoryear{Feng and Boyd-Graber}{Feng and
  Boyd-Graber}{2019}]%
        {whatcanAIdoforme}
\bibfield{author}{\bibinfo{person}{Shi Feng} {and} \bibinfo{person}{Jordan
  Boyd-Graber}.} \bibinfo{year}{2019}\natexlab{}.
\newblock \showarticletitle{What Can AI Do for Me? Evaluating Machine Learning
  Interpretations in Cooperative Play}. In
  \bibinfo{booktitle}{\emph{Proceedings of the 24th International Conference on
  Intelligent User Interfaces}} \emph{(\bibinfo{series}{IUI ’19})}.
  \bibinfo{publisher}{Association for Computing Machinery},
  \bibinfo{address}{New York, NY, USA}, \bibinfo{pages}{229–239}.
\newblock
\showISBNx{9781450362726}
\urldef\tempurl%
\url{https://doi.org/10.1145/3301275.3302265}
\showDOI{\tempurl}


\bibitem[\protect\citeauthoryear{Ghahramani and Jordan}{Ghahramani and
  Jordan}{1996}]%
        {ghahramani1996factorial}
\bibfield{author}{\bibinfo{person}{Zoubin Ghahramani} {and}
  \bibinfo{person}{Michael~I Jordan}.} \bibinfo{year}{1996}\natexlab{}.
\newblock \showarticletitle{Factorial hidden Markov models}. In
  \bibinfo{booktitle}{\emph{Advances in Neural Information Processing
  Systems}}. \bibinfo{pages}{472--478}.
\newblock


\bibitem[\protect\citeauthoryear{Gilpin, Bau, Yuan, Bajwa, Specter, and
  Kagal}{Gilpin et~al\mbox{.}}{2018}]%
        {gilpin2018explaining}
\bibfield{author}{\bibinfo{person}{Leilani~H Gilpin}, \bibinfo{person}{David
  Bau}, \bibinfo{person}{Ben~Z Yuan}, \bibinfo{person}{Ayesha Bajwa},
  \bibinfo{person}{Michael Specter}, {and} \bibinfo{person}{Lalana Kagal}.}
  \bibinfo{year}{2018}\natexlab{}.
\newblock \showarticletitle{Explaining explanations: An overview of
  interpretability of machine learning}. In \bibinfo{booktitle}{\emph{2018 IEEE
  5th International Conference on data science and advanced analytics (DSAA)}}.
  IEEE, \bibinfo{pages}{80--89}.
\newblock


\bibitem[\protect\citeauthoryear{Goodfellow, Pouget-Abadie, Mirza, Xu,
  Warde-Farley, Ozair, Courville, and Bengio}{Goodfellow et~al\mbox{.}}{2014}]%
        {goodfellow2014generative}
\bibfield{author}{\bibinfo{person}{Ian Goodfellow}, \bibinfo{person}{Jean
  Pouget-Abadie}, \bibinfo{person}{Mehdi Mirza}, \bibinfo{person}{Bing Xu},
  \bibinfo{person}{David Warde-Farley}, \bibinfo{person}{Sherjil Ozair},
  \bibinfo{person}{Aaron Courville}, {and} \bibinfo{person}{Yoshua Bengio}.}
  \bibinfo{year}{2014}\natexlab{}.
\newblock \showarticletitle{Generative adversarial nets}. In
  \bibinfo{booktitle}{\emph{Advances in neural information processing
  systems}}. \bibinfo{pages}{2672--2680}.
\newblock


\bibitem[\protect\citeauthoryear{Ha and Schmidhuber}{Ha and
  Schmidhuber}{2018}]%
        {ha2018worldmodels}
\bibfield{author}{\bibinfo{person}{David Ha} {and} \bibinfo{person}{J{\"u}rgen
  Schmidhuber}.} \bibinfo{year}{2018}\natexlab{}.
\newblock \showarticletitle{Recurrent World Models Facilitate Policy
  Evolution}.
\newblock In \bibinfo{booktitle}{\emph{Advances in Neural Information
  Processing Systems 31}}. \bibinfo{publisher}{Curran Associates, Inc.},
  \bibinfo{pages}{2451--2463}.
\newblock
\urldef\tempurl%
\url{https://papers.nips.cc/paper/7512-recurrent-world-models-facilitate-policy-evolution}
\showURL{%
\tempurl}
\newblock
\shownote{\url{https://worldmodels.github.io}.}


\bibitem[\protect\citeauthoryear{Hansen, Miron-Shatz, Lau, and Paton}{Hansen
  et~al\mbox{.}}{2014}]%
        {hansen2014big}
\bibfield{author}{\bibinfo{person}{MM Hansen}, \bibinfo{person}{T Miron-Shatz},
  \bibinfo{person}{AYS Lau}, {and} \bibinfo{person}{C Paton}.}
  \bibinfo{year}{2014}\natexlab{}.
\newblock \showarticletitle{Big data in science and healthcare: a review of
  recent literature and perspectives}.
\newblock \bibinfo{journal}{\emph{Yearbook of medical informatics}}
  \bibinfo{volume}{23}, \bibinfo{number}{01} (\bibinfo{year}{2014}),
  \bibinfo{pages}{21--26}.
\newblock


\bibitem[\protect\citeauthoryear{Hart and Staveland}{Hart and
  Staveland}{1988}]%
        {hart1988development}
\bibfield{author}{\bibinfo{person}{Sandra~G Hart} {and}
  \bibinfo{person}{Lowell~E Staveland}.} \bibinfo{year}{1988}\natexlab{}.
\newblock \showarticletitle{Development of NASA-TLX (Task Load Index): Results
  of empirical and theoretical research}.
\newblock In \bibinfo{booktitle}{\emph{Advances in psychology}}.
  Vol.~\bibinfo{volume}{52}. \bibinfo{publisher}{Elsevier},
  \bibinfo{pages}{139--183}.
\newblock


\bibitem[\protect\citeauthoryear{Higgins, Amos, Pfau, Racaniere, Matthey,
  Rezende, and Lerchner}{Higgins et~al\mbox{.}}{2018}]%
        {higgins2018towards}
\bibfield{author}{\bibinfo{person}{Irina Higgins}, \bibinfo{person}{David
  Amos}, \bibinfo{person}{David Pfau}, \bibinfo{person}{Sebastien Racaniere},
  \bibinfo{person}{Loic Matthey}, \bibinfo{person}{Danilo Rezende}, {and}
  \bibinfo{person}{Alexander Lerchner}.} \bibinfo{year}{2018}\natexlab{}.
\newblock \showarticletitle{Towards a definition of disentangled
  representations}.
\newblock \bibinfo{journal}{\emph{arXiv preprint arXiv:1812.02230}}
  (\bibinfo{year}{2018}).
\newblock


\bibitem[\protect\citeauthoryear{Higgins, Matthey, Pal, Burgess, Glorot,
  Botvinick, Mohamed, and Lerchner}{Higgins et~al\mbox{.}}{2017}]%
        {higgins2017beta}
\bibfield{author}{\bibinfo{person}{Irina Higgins}, \bibinfo{person}{Loic
  Matthey}, \bibinfo{person}{Arka Pal}, \bibinfo{person}{Christopher Burgess},
  \bibinfo{person}{Xavier Glorot}, \bibinfo{person}{Matthew Botvinick},
  \bibinfo{person}{Shakir Mohamed}, {and} \bibinfo{person}{Alexander
  Lerchner}.} \bibinfo{year}{2017}\natexlab{}.
\newblock \showarticletitle{beta-vae: Learning basic visual concepts with a
  constrained variational framework}. In
  \bibinfo{booktitle}{\emph{International Conference on Learning
  Representations}}, Vol.~\bibinfo{volume}{3}.
\newblock


\bibitem[\protect\citeauthoryear{Hinton and Salakhutdinov}{Hinton and
  Salakhutdinov}{2006}]%
        {hinton2006reducing}
\bibfield{author}{\bibinfo{person}{Geoffrey~E Hinton} {and}
  \bibinfo{person}{Ruslan~R Salakhutdinov}.} \bibinfo{year}{2006}\natexlab{}.
\newblock \showarticletitle{Reducing the dimensionality of data with neural
  networks}.
\newblock \bibinfo{journal}{\emph{science}} \bibinfo{volume}{313},
  \bibinfo{number}{5786} (\bibinfo{year}{2006}), \bibinfo{pages}{504--507}.
\newblock


\bibitem[\protect\citeauthoryear{Huysmans, Dejaeger, Mues, Vanthienen, and
  Baesens}{Huysmans et~al\mbox{.}}{2011}]%
        {huysmans2011empirical}
\bibfield{author}{\bibinfo{person}{Johan Huysmans}, \bibinfo{person}{Karel
  Dejaeger}, \bibinfo{person}{Christophe Mues}, \bibinfo{person}{Jan
  Vanthienen}, {and} \bibinfo{person}{Bart Baesens}.}
  \bibinfo{year}{2011}\natexlab{}.
\newblock \showarticletitle{An empirical evaluation of the comprehensibility of
  decision table, tree and rule based predictive models}.
\newblock \bibinfo{journal}{\emph{Decision Support Systems}}
  \bibinfo{volume}{51}, \bibinfo{number}{1} (\bibinfo{year}{2011}),
  \bibinfo{pages}{141--154}.
\newblock


\bibitem[\protect\citeauthoryear{Jolliffe}{Jolliffe}{1986}]%
        {jolliffe1986principal}
\bibfield{author}{\bibinfo{person}{Ian~T Jolliffe}.}
  \bibinfo{year}{1986}\natexlab{}.
\newblock \showarticletitle{Principal components in regression analysis}.
\newblock In \bibinfo{booktitle}{\emph{Principal component analysis}}.
  \bibinfo{publisher}{Springer}, \bibinfo{pages}{129--155}.
\newblock


\bibitem[\protect\citeauthoryear{Kaur, Nori, Jenkins, Caruana, Wallach, and
  Wortman~Vaughan}{Kaur et~al\mbox{.}}{2020}]%
        {InterpretingInterpretability}
\bibfield{author}{\bibinfo{person}{Harmanpreet Kaur}, \bibinfo{person}{Harsha
  Nori}, \bibinfo{person}{Samuel Jenkins}, \bibinfo{person}{Rich Caruana},
  \bibinfo{person}{Hanna Wallach}, {and} \bibinfo{person}{Jennifer
  Wortman~Vaughan}.} \bibinfo{year}{2020}\natexlab{}.
\newblock \showarticletitle{Interpreting Interpretability: Understanding Data
  Scientists’ Use of Interpretability Tools for Machine Learning}. In
  \bibinfo{booktitle}{\emph{Proceedings of the 2020 CHI Conference on Human
  Factors in Computing Systems}} \emph{(\bibinfo{series}{CHI ’20})}.
  \bibinfo{publisher}{Association for Computing Machinery},
  \bibinfo{address}{New York, NY, USA}, \bibinfo{pages}{1–14}.
\newblock
\showISBNx{9781450367080}
\urldef\tempurl%
\url{https://doi.org/10.1145/3313831.3376219}
\showDOI{\tempurl}


\bibitem[\protect\citeauthoryear{Kim, Wattenberg, Gilmer, Cai, Wexler, Viegas,
  et~al\mbox{.}}{Kim et~al\mbox{.}}{2018}]%
        {kim2018interpretability}
\bibfield{author}{\bibinfo{person}{Been Kim}, \bibinfo{person}{Martin
  Wattenberg}, \bibinfo{person}{Justin Gilmer}, \bibinfo{person}{Carrie Cai},
  \bibinfo{person}{James Wexler}, \bibinfo{person}{Fernanda Viegas},
  {et~al\mbox{.}}} \bibinfo{year}{2018}\natexlab{}.
\newblock \showarticletitle{Interpretability beyond feature attribution:
  Quantitative testing with concept activation vectors (tcav)}. In
  \bibinfo{booktitle}{\emph{International conference on machine learning}}.
  PMLR, \bibinfo{pages}{2668--2677}.
\newblock


\bibitem[\protect\citeauthoryear{Kim and Mnih}{Kim and Mnih}{2018}]%
        {kim2018disentangling}
\bibfield{author}{\bibinfo{person}{Hyunjik Kim} {and} \bibinfo{person}{Andriy
  Mnih}.} \bibinfo{year}{2018}\natexlab{}.
\newblock \showarticletitle{Disentangling by Factorising}. In
  \bibinfo{booktitle}{\emph{Proceedings of the 35th International Conference on
  Machine Learning}} \emph{(\bibinfo{series}{Proceedings of Machine Learning
  Research})}, \bibfield{editor}{\bibinfo{person}{Jennifer Dy} {and}
  \bibinfo{person}{Andreas Krause}} (Eds.), Vol.~\bibinfo{volume}{80}.
  \bibinfo{publisher}{PMLR}, \bibinfo{address}{Stockholmsmässan, Stockholm
  Sweden}, \bibinfo{pages}{2649--2658}.
\newblock
\urldef\tempurl%
\url{http://proceedings.mlr.press/v80/kim18b.html}
\showURL{%
\tempurl}


\bibitem[\protect\citeauthoryear{Kingma and Ba}{Kingma and Ba}{2014}]%
        {kingma2014adam}
\bibfield{author}{\bibinfo{person}{Diederik~P Kingma} {and}
  \bibinfo{person}{Jimmy Ba}.} \bibinfo{year}{2014}\natexlab{}.
\newblock \showarticletitle{Adam: A method for stochastic optimization}.
\newblock \bibinfo{journal}{\emph{arXiv preprint arXiv:1412.6980}}
  (\bibinfo{year}{2014}).
\newblock


\bibitem[\protect\citeauthoryear{Kingma and Welling}{Kingma and
  Welling}{2013}]%
        {kingma2013auto}
\bibfield{author}{\bibinfo{person}{Diederik~P Kingma} {and}
  \bibinfo{person}{Max Welling}.} \bibinfo{year}{2013}\natexlab{}.
\newblock \showarticletitle{Auto-encoding variational bayes}.
\newblock \bibinfo{journal}{\emph{arXiv preprint arXiv:1312.6114}}
  (\bibinfo{year}{2013}).
\newblock


\bibitem[\protect\citeauthoryear{Krause, Perer, and Ng}{Krause
  et~al\mbox{.}}{2016a}]%
        {krause2016interacting}
\bibfield{author}{\bibinfo{person}{Josua Krause}, \bibinfo{person}{Adam Perer},
  {and} \bibinfo{person}{Kenney Ng}.} \bibinfo{year}{2016}\natexlab{a}.
\newblock \showarticletitle{Interacting with predictions: Visual inspection of
  black-box machine learning models}. In \bibinfo{booktitle}{\emph{Proceedings
  of the 2016 CHI Conference on Human Factors in Computing Systems}}.
  \bibinfo{pages}{5686--5697}.
\newblock


\bibitem[\protect\citeauthoryear{Krause, Perer, and Ng}{Krause
  et~al\mbox{.}}{2016b}]%
        {interacting-with-predictions}
\bibfield{author}{\bibinfo{person}{Josua Krause}, \bibinfo{person}{Adam Perer},
  {and} \bibinfo{person}{Kenney Ng}.} \bibinfo{year}{2016}\natexlab{b}.
\newblock \showarticletitle{Interacting with Predictions: Visual Inspection of
  Black-Box Machine Learning Models}. In \bibinfo{booktitle}{\emph{Proceedings
  of the 2016 CHI Conference on Human Factors in Computing Systems}}
  \emph{(\bibinfo{series}{CHI '16})}. \bibinfo{publisher}{Association for
  Computing Machinery}, \bibinfo{address}{New York, NY, USA},
  \bibinfo{pages}{5686–5697}.
\newblock
\showISBNx{9781450333627}
\urldef\tempurl%
\url{https://doi.org/10.1145/2858036.2858529}
\showDOI{\tempurl}


\bibitem[\protect\citeauthoryear{Kulesza, Stumpf, Burnett, Yang, Kwan, and
  Wong}{Kulesza et~al\mbox{.}}{2013}]%
        {kulesza2013too}
\bibfield{author}{\bibinfo{person}{Todd Kulesza}, \bibinfo{person}{Simone
  Stumpf}, \bibinfo{person}{Margaret Burnett}, \bibinfo{person}{Sherry Yang},
  \bibinfo{person}{Irwin Kwan}, {and} \bibinfo{person}{Weng-Keen Wong}.}
  \bibinfo{year}{2013}\natexlab{}.
\newblock \showarticletitle{Too much, too little, or just right? Ways
  explanations impact end users' mental models}. In
  \bibinfo{booktitle}{\emph{2013 IEEE Symposium on Visual Languages and Human
  Centric Computing}}. IEEE, \bibinfo{pages}{3--10}.
\newblock


\bibitem[\protect\citeauthoryear{Lage, Chen, He, Narayanan, Kim, Gershman, and
  Doshi-Velez}{Lage et~al\mbox{.}}{2019}]%
        {lage2019evaluation}
\bibfield{author}{\bibinfo{person}{Isaac Lage}, \bibinfo{person}{Emily Chen},
  \bibinfo{person}{Jeffrey He}, \bibinfo{person}{Menaka Narayanan},
  \bibinfo{person}{Been Kim}, \bibinfo{person}{Sam Gershman}, {and}
  \bibinfo{person}{Finale Doshi-Velez}.} \bibinfo{year}{2019}\natexlab{}.
\newblock \showarticletitle{An evaluation of the human-interpretability of
  explanation}.
\newblock \bibinfo{journal}{\emph{arXiv preprint arXiv:1902.00006}}
  (\bibinfo{year}{2019}).
\newblock


\bibitem[\protect\citeauthoryear{Lage, Ross, Gershman, Kim, and
  Doshi-Velez}{Lage et~al\mbox{.}}{2018}]%
        {lage2018human}
\bibfield{author}{\bibinfo{person}{Isaac Lage}, \bibinfo{person}{Andrew Ross},
  \bibinfo{person}{Samuel~J Gershman}, \bibinfo{person}{Been Kim}, {and}
  \bibinfo{person}{Finale Doshi-Velez}.} \bibinfo{year}{2018}\natexlab{}.
\newblock \showarticletitle{Human-in-the-loop interpretability prior}. In
  \bibinfo{booktitle}{\emph{Advances in neural information processing
  systems}}. \bibinfo{pages}{10159--10168}.
\newblock


\bibitem[\protect\citeauthoryear{LeCun}{LeCun}{1998}]%
        {lecun1998mnist}
\bibfield{author}{\bibinfo{person}{Yann LeCun}.}
  \bibinfo{year}{1998}\natexlab{}.
\newblock \showarticletitle{The MNIST database of handwritten digits}.
\newblock \bibinfo{journal}{\emph{http://yann. lecun. com/exdb/mnist/}}
  (\bibinfo{year}{1998}).
\newblock


\bibitem[\protect\citeauthoryear{Levandowsky and Winter}{Levandowsky and
  Winter}{1971}]%
        {levandowsky1971distance}
\bibfield{author}{\bibinfo{person}{Michael Levandowsky} {and}
  \bibinfo{person}{David Winter}.} \bibinfo{year}{1971}\natexlab{}.
\newblock \showarticletitle{Distance between sets}.
\newblock \bibinfo{journal}{\emph{Nature}} \bibinfo{volume}{234},
  \bibinfo{number}{5323} (\bibinfo{year}{1971}), \bibinfo{pages}{34--35}.
\newblock


\bibitem[\protect\citeauthoryear{Lewis}{Lewis}{1982}]%
        {lewis1982using}
\bibfield{author}{\bibinfo{person}{Clayton Lewis}.}
  \bibinfo{year}{1982}\natexlab{}.
\newblock \bibinfo{booktitle}{\emph{Using the" thinking-aloud" method in
  cognitive interface design}}.
\newblock \bibinfo{publisher}{IBM TJ Watson Research Center Yorktown Heights,
  NY}.
\newblock


\bibitem[\protect\citeauthoryear{Lim, Dey, and Avrahami}{Lim
  et~al\mbox{.}}{2009}]%
        {lim2009and}
\bibfield{author}{\bibinfo{person}{Brian~Y Lim}, \bibinfo{person}{Anind~K Dey},
  {and} \bibinfo{person}{Daniel Avrahami}.} \bibinfo{year}{2009}\natexlab{}.
\newblock \showarticletitle{Why and why not explanations improve the
  intelligibility of context-aware intelligent systems}. In
  \bibinfo{booktitle}{\emph{Proceedings of the SIGCHI Conference on Human
  Factors in Computing Systems}}. \bibinfo{pages}{2119--2128}.
\newblock


\bibitem[\protect\citeauthoryear{Lipton}{Lipton}{2018}]%
        {lipton2018mythos}
\bibfield{author}{\bibinfo{person}{Zachary~C Lipton}.}
  \bibinfo{year}{2018}\natexlab{}.
\newblock \showarticletitle{The mythos of model interpretability}.
\newblock \bibinfo{journal}{\emph{Queue}} \bibinfo{volume}{16},
  \bibinfo{number}{3} (\bibinfo{year}{2018}), \bibinfo{pages}{31--57}.
\newblock


\bibitem[\protect\citeauthoryear{Locatello, Bauer, Lucic, Raetsch, Gelly,
  Sch{\"o}lkopf, and Bachem}{Locatello et~al\mbox{.}}{2019}]%
        {locatello2018challenging}
\bibfield{author}{\bibinfo{person}{Francesco Locatello},
  \bibinfo{person}{Stefan Bauer}, \bibinfo{person}{Mario Lucic},
  \bibinfo{person}{Gunnar Raetsch}, \bibinfo{person}{Sylvain Gelly},
  \bibinfo{person}{Bernhard Sch{\"o}lkopf}, {and} \bibinfo{person}{Olivier
  Bachem}.} \bibinfo{year}{2019}\natexlab{}.
\newblock \showarticletitle{Challenging Common Assumptions in the Unsupervised
  Learning of Disentangled Representations}. In
  \bibinfo{booktitle}{\emph{Proceedings of the 36th International Conference on
  Machine Learning}} \emph{(\bibinfo{series}{Proceedings of Machine Learning
  Research})}, \bibfield{editor}{\bibinfo{person}{Kamalika Chaudhuri} {and}
  \bibinfo{person}{Ruslan Salakhutdinov}} (Eds.), Vol.~\bibinfo{volume}{97}.
  \bibinfo{publisher}{PMLR}, \bibinfo{address}{Long Beach, California, USA},
  \bibinfo{pages}{4114--4124}.
\newblock
\urldef\tempurl%
\url{http://proceedings.mlr.press/v97/locatello19a.html}
\showURL{%
\tempurl}


\bibitem[\protect\citeauthoryear{Maaten and Hinton}{Maaten and Hinton}{2008}]%
        {maaten2008visualizing}
\bibfield{author}{\bibinfo{person}{Laurens van~der Maaten} {and}
  \bibinfo{person}{Geoffrey Hinton}.} \bibinfo{year}{2008}\natexlab{}.
\newblock \showarticletitle{Visualizing data using t-SNE}.
\newblock \bibinfo{journal}{\emph{Journal of machine learning research}}
  \bibinfo{volume}{9}, \bibinfo{number}{Nov} (\bibinfo{year}{2008}),
  \bibinfo{pages}{2579--2605}.
\newblock


\bibitem[\protect\citeauthoryear{Matthey, Higgins, Hassabis, and
  Lerchner}{Matthey et~al\mbox{.}}{2017}]%
        {dsprites17}
\bibfield{author}{\bibinfo{person}{Loic Matthey}, \bibinfo{person}{Irina
  Higgins}, \bibinfo{person}{Demis Hassabis}, {and} \bibinfo{person}{Alexander
  Lerchner}.} \bibinfo{year}{2017}\natexlab{}.
\newblock \bibinfo{title}{dSprites: Disentanglement testing Sprites dataset}.
\newblock
  \bibinfo{howpublished}{\url{https://github.com/deepmind/dsprites-dataset}}.
\newblock


\bibitem[\protect\citeauthoryear{Miller}{Miller}{2019}]%
        {miller2019explanation}
\bibfield{author}{\bibinfo{person}{Tim Miller}.}
  \bibinfo{year}{2019}\natexlab{}.
\newblock \showarticletitle{Explanation in artificial intelligence: Insights
  from the social sciences}.
\newblock \bibinfo{journal}{\emph{Artificial Intelligence}}
  \bibinfo{volume}{267} (\bibinfo{year}{2019}), \bibinfo{pages}{1--38}.
\newblock


\bibitem[\protect\citeauthoryear{Norman}{Norman}{2013}]%
        {norman2013design}
\bibfield{author}{\bibinfo{person}{Don Norman}.}
  \bibinfo{year}{2013}\natexlab{}.
\newblock \bibinfo{booktitle}{\emph{The design of everyday things: Revised and
  expanded edition}}.
\newblock \bibinfo{publisher}{Basic books}.
\newblock


\bibitem[\protect\citeauthoryear{Olah, Mordvintsev, and Schubert}{Olah
  et~al\mbox{.}}{2017}]%
        {olah2017feature}
\bibfield{author}{\bibinfo{person}{Chris Olah}, \bibinfo{person}{Alexander
  Mordvintsev}, {and} \bibinfo{person}{Ludwig Schubert}.}
  \bibinfo{year}{2017}\natexlab{}.
\newblock \showarticletitle{Feature visualization}.
\newblock \bibinfo{journal}{\emph{Distill}} \bibinfo{volume}{2},
  \bibinfo{number}{11} (\bibinfo{year}{2017}), \bibinfo{pages}{e7}.
\newblock


\bibitem[\protect\citeauthoryear{Olah, Satyanarayan, Johnson, Carter, Schubert,
  Ye, and Mordvintsev}{Olah et~al\mbox{.}}{2018}]%
        {olah2018building}
\bibfield{author}{\bibinfo{person}{Chris Olah}, \bibinfo{person}{Arvind
  Satyanarayan}, \bibinfo{person}{Ian Johnson}, \bibinfo{person}{Shan Carter},
  \bibinfo{person}{Ludwig Schubert}, \bibinfo{person}{Katherine Ye}, {and}
  \bibinfo{person}{Alexander Mordvintsev}.} \bibinfo{year}{2018}\natexlab{}.
\newblock \showarticletitle{The building blocks of interpretability}.
\newblock \bibinfo{journal}{\emph{Distill}} \bibinfo{volume}{3},
  \bibinfo{number}{3} (\bibinfo{year}{2018}), \bibinfo{pages}{e10}.
\newblock


\bibitem[\protect\citeauthoryear{Poursabzi-Sangdeh, Goldstein, Hofman, Vaughan,
  and Wallach}{Poursabzi-Sangdeh et~al\mbox{.}}{2018}]%
        {poursabzi2018manipulating}
\bibfield{author}{\bibinfo{person}{Forough Poursabzi-Sangdeh},
  \bibinfo{person}{Daniel~G Goldstein}, \bibinfo{person}{Jake~M Hofman},
  \bibinfo{person}{Jennifer~Wortman Vaughan}, {and} \bibinfo{person}{Hanna
  Wallach}.} \bibinfo{year}{2018}\natexlab{}.
\newblock \showarticletitle{Manipulating and measuring model interpretability}.
\newblock \bibinfo{journal}{\emph{arXiv preprint arXiv:1802.07810}}
  (\bibinfo{year}{2018}).
\newblock


\bibitem[\protect\citeauthoryear{Ridgeway}{Ridgeway}{2016}]%
        {ridgeway2016survey}
\bibfield{author}{\bibinfo{person}{Karl Ridgeway}.}
  \bibinfo{year}{2016}\natexlab{}.
\newblock \showarticletitle{A survey of inductive biases for factorial
  representation-learning}.
\newblock \bibinfo{journal}{\emph{arXiv preprint arXiv:1612.05299}}
  (\bibinfo{year}{2016}).
\newblock


\bibitem[\protect\citeauthoryear{Rozenblit and Keil}{Rozenblit and
  Keil}{2002}]%
        {rozenblit2002misunderstood}
\bibfield{author}{\bibinfo{person}{Leonid Rozenblit} {and}
  \bibinfo{person}{Frank Keil}.} \bibinfo{year}{2002}\natexlab{}.
\newblock \showarticletitle{The misunderstood limits of folk science: An
  illusion of explanatory depth}.
\newblock \bibinfo{journal}{\emph{Cognitive science}} \bibinfo{volume}{26},
  \bibinfo{number}{5} (\bibinfo{year}{2002}), \bibinfo{pages}{521--562}.
\newblock


\bibitem[\protect\citeauthoryear{Sauro and Dumas}{Sauro and Dumas}{2009}]%
        {sauro2009comparison}
\bibfield{author}{\bibinfo{person}{Jeff Sauro} {and} \bibinfo{person}{Joseph~S
  Dumas}.} \bibinfo{year}{2009}\natexlab{}.
\newblock \showarticletitle{Comparison of three one-question, post-task
  usability questionnaires}. In \bibinfo{booktitle}{\emph{Proceedings of the
  SIGCHI conference on human factors in computing systems}}.
  \bibinfo{pages}{1599--1608}.
\newblock


\bibitem[\protect\citeauthoryear{Scheuerman, Spiel, Haimson, Hamidi, and
  Branham}{Scheuerman et~al\mbox{.}}{2019}]%
        {scheuerman2019hci}
\bibfield{author}{\bibinfo{person}{Morgan~Klaus Scheuerman},
  \bibinfo{person}{Katta Spiel}, \bibinfo{person}{Oliver~L Haimson},
  \bibinfo{person}{Foad Hamidi}, {and} \bibinfo{person}{Stacy~M Branham}.}
  \bibinfo{year}{2019}\natexlab{}.
\newblock \bibinfo{title}{HCI guidelines for gender equity and inclusivity}.
\newblock
  \bibinfo{howpublished}{\url{https://www.morgan-klaus.com/gender-guidelines.html}}.
\newblock


\bibitem[\protect\citeauthoryear{Schmid, Zeller, Besold, Tamaddoni-Nezhad, and
  Muggleton}{Schmid et~al\mbox{.}}{2016}]%
        {schmid2016does}
\bibfield{author}{\bibinfo{person}{Ute Schmid}, \bibinfo{person}{Christina
  Zeller}, \bibinfo{person}{Tarek Besold}, \bibinfo{person}{Alireza
  Tamaddoni-Nezhad}, {and} \bibinfo{person}{Stephen Muggleton}.}
  \bibinfo{year}{2016}\natexlab{}.
\newblock \showarticletitle{How does predicate invention affect human
  comprehensibility?}. In \bibinfo{booktitle}{\emph{International Conference on
  Inductive Logic Programming}}. Springer, \bibinfo{pages}{52--67}.
\newblock


\bibitem[\protect\citeauthoryear{Sepliarskaia, Kiseleva, and
  de~Rijke}{Sepliarskaia et~al\mbox{.}}{2019}]%
        {sepliarskaia2019evaluating}
\bibfield{author}{\bibinfo{person}{Anna Sepliarskaia}, \bibinfo{person}{Julia
  Kiseleva}, {and} \bibinfo{person}{Maarten de Rijke}.}
  \bibinfo{year}{2019}\natexlab{}.
\newblock \showarticletitle{Evaluating Disentangled Representations}.
\newblock \bibinfo{journal}{\emph{arXiv preprint arXiv:1910.05587}}
  (\bibinfo{year}{2019}).
\newblock


\bibitem[\protect\citeauthoryear{Slack, Friedler, Scheidegger, and Roy}{Slack
  et~al\mbox{.}}{2019}]%
        {slack2019assessing}
\bibfield{author}{\bibinfo{person}{Dylan Slack}, \bibinfo{person}{Sorelle~A
  Friedler}, \bibinfo{person}{Carlos Scheidegger}, {and}
  \bibinfo{person}{Chitradeep~Dutta Roy}.} \bibinfo{year}{2019}\natexlab{}.
\newblock \showarticletitle{Assessing the Local Interpretability of Machine
  Learning Models}.
\newblock \bibinfo{journal}{\emph{arXiv preprint arXiv:1902.03501}}
  (\bibinfo{year}{2019}).
\newblock


\bibitem[\protect\citeauthoryear{Smilkov, Thorat, Assogba, Yuan, Kreeger, Yu,
  Zhang, Cai, Nielsen, Soergel, et~al\mbox{.}}{Smilkov et~al\mbox{.}}{2019}]%
        {smilkov2019tensorflow}
\bibfield{author}{\bibinfo{person}{Daniel Smilkov}, \bibinfo{person}{Nikhil
  Thorat}, \bibinfo{person}{Yannick Assogba}, \bibinfo{person}{Ann Yuan},
  \bibinfo{person}{Nick Kreeger}, \bibinfo{person}{Ping Yu},
  \bibinfo{person}{Kangyi Zhang}, \bibinfo{person}{Shanqing Cai},
  \bibinfo{person}{Eric Nielsen}, \bibinfo{person}{David Soergel},
  {et~al\mbox{.}}} \bibinfo{year}{2019}\natexlab{}.
\newblock \showarticletitle{Tensorflow. js: Machine learning for the web and
  beyond}.
\newblock \bibinfo{journal}{\emph{arXiv preprint arXiv:1901.05350}}
  (\bibinfo{year}{2019}).
\newblock


\bibitem[\protect\citeauthoryear{Smilkov, Thorat, Nicholson, Reif, Vi{\'e}gas,
  and Wattenberg}{Smilkov et~al\mbox{.}}{2016}]%
        {smilkov2016embedding}
\bibfield{author}{\bibinfo{person}{Daniel Smilkov}, \bibinfo{person}{Nikhil
  Thorat}, \bibinfo{person}{Charles Nicholson}, \bibinfo{person}{Emily Reif},
  \bibinfo{person}{Fernanda~B Vi{\'e}gas}, {and} \bibinfo{person}{Martin
  Wattenberg}.} \bibinfo{year}{2016}\natexlab{}.
\newblock \showarticletitle{Embedding projector: Interactive visualization and
  interpretation of embeddings}.
\newblock \bibinfo{journal}{\emph{arXiv preprint arXiv:1611.05469}}
  (\bibinfo{year}{2016}).
\newblock


\bibitem[\protect\citeauthoryear{Sweller}{Sweller}{1994}]%
        {sweller1994cognitive}
\bibfield{author}{\bibinfo{person}{John Sweller}.}
  \bibinfo{year}{1994}\natexlab{}.
\newblock \showarticletitle{Cognitive load theory, learning difficulty, and
  instructional design}.
\newblock \bibinfo{journal}{\emph{Learning and instruction}}
  \bibinfo{volume}{4}, \bibinfo{number}{4} (\bibinfo{year}{1994}),
  \bibinfo{pages}{295--312}.
\newblock


\bibitem[\protect\citeauthoryear{Wang, Bovik, Sheikh, and Simoncelli}{Wang
  et~al\mbox{.}}{2004}]%
        {wang2004image}
\bibfield{author}{\bibinfo{person}{Zhou Wang}, \bibinfo{person}{Alan~C Bovik},
  \bibinfo{person}{Hamid~R Sheikh}, {and} \bibinfo{person}{Eero~P Simoncelli}.}
  \bibinfo{year}{2004}\natexlab{}.
\newblock \showarticletitle{Image quality assessment: from error visibility to
  structural similarity}.
\newblock \bibinfo{journal}{\emph{IEEE transactions on image processing}}
  \bibinfo{volume}{13}, \bibinfo{number}{4} (\bibinfo{year}{2004}),
  \bibinfo{pages}{600--612}.
\newblock


\bibitem[\protect\citeauthoryear{Wexler, Pushkarna, Bolukbasi, Wattenberg,
  Vi{\'e}gas, and Wilson}{Wexler et~al\mbox{.}}{2019}]%
        {wexler2019if}
\bibfield{author}{\bibinfo{person}{James Wexler}, \bibinfo{person}{Mahima
  Pushkarna}, \bibinfo{person}{Tolga Bolukbasi}, \bibinfo{person}{Martin
  Wattenberg}, \bibinfo{person}{Fernanda Vi{\'e}gas}, {and}
  \bibinfo{person}{Jimbo Wilson}.} \bibinfo{year}{2019}\natexlab{}.
\newblock \showarticletitle{The what-if tool: Interactive probing of machine
  learning models}.
\newblock \bibinfo{journal}{\emph{IEEE transactions on visualization and
  computer graphics}} \bibinfo{volume}{26}, \bibinfo{number}{1}
  (\bibinfo{year}{2019}), \bibinfo{pages}{56--65}.
\newblock


\bibitem[\protect\citeauthoryear{Wu et~al\mbox{.}}{Wu et~al\mbox{.}}{2016}]%
        {wu2016tensorpack}
\bibfield{author}{\bibinfo{person}{Yuxin Wu} {et~al\mbox{.}}}
  \bibinfo{year}{2016}\natexlab{}.
\newblock \bibinfo{title}{Tensorpack}.
\newblock \bibinfo{howpublished}{\url{https://github.com/tensorpack/}}.
\newblock


\bibitem[\protect\citeauthoryear{Zhang, Isola, Efros, Shechtman, and
  Wang}{Zhang et~al\mbox{.}}{2018}]%
        {zhang2018unreasonable}
\bibfield{author}{\bibinfo{person}{Richard Zhang}, \bibinfo{person}{Phillip
  Isola}, \bibinfo{person}{Alexei~A Efros}, \bibinfo{person}{Eli Shechtman},
  {and} \bibinfo{person}{Oliver Wang}.} \bibinfo{year}{2018}\natexlab{}.
\newblock \showarticletitle{The unreasonable effectiveness of deep features as
  a perceptual metric}. In \bibinfo{booktitle}{\emph{Proceedings of the IEEE
  conference on computer vision and pattern recognition}}.
  \bibinfo{pages}{586--595}.
\newblock


\end{thebibliography}

\appendix
\renewcommand{\thesection}{A}
\renewcommand\thefigure{\thesection.\arabic{figure}}
\renewcommand\thetable{\thesection.\arabic{table}}

\section{Appendix}

\subsection{Additional Model Details}
\label{sec:arch-details}

In this section, we present additional background and details about the representation learning models considered.

\subsubsection{Loss Functions.}
Autoencoders \cite{ackley1985learning} (AEs) are trained simply to compress and reconstruct $x$: \[
    \mathcal{L}_{\mathrm{AE}}(x) = \mathcal{L}(x, \mathtt{dec}(\mathtt{enc}(x))),
\]
where $\mathcal{L}$ is an individual example reconstruction loss. We used cross-entropy (Bernoulli negative log-likelihood) for dSprites and MNIST and mean-squared error (Gaussian negative log-likelihood) for Sinelines.

Ground-truth models (GTs) are trained equivalently to the autoencoder, except we omit the encoder and instead provide ground-truth factors $z$ alongside $x$: \[
    \mathcal{L}_{\mathrm{GT}}(x,z) = \mathcal{L}(x, \mathtt{dec}(z))
\]
Note that on Sinelines, we did not actually \emph{train} the GT model because in that case, it was simple to implement in closed form. It might have been possible to implement the GT dSprites model this way as well, but since the original generating code is not available in the dataset repository \cite{dsprites17}, we opted for a model.

Variational autoencoders \cite{kingma2013auto} (VAEs) are similar to autoencoders, except that $\enc(x)$ outputs not a single value, but a distribution over $z$. The VAE training objective includes an expectation of the reconstruction error over this distribution, as well as a regularization term meant to ensure that $\enc(x)$ stays close to a prior $p(z)$ (in our case, an isotropic unit Gaussian): \[
\mathcal{L}_{\mathrm{VAE}}(x) = \mathbb{E}_{z \sim \mathtt{enc}(x)} \left[ \mathcal{L}(x, \mathtt{dec}(z)) \right] + \mathrm{KL}(\mathtt{enc}(x)||p(z))
\]
Note that the expectation is generally approximated in training with a single sample.

$\beta$-total correlation autoencoders ($\beta$-TCVAEs, abbreviated further to TC) \cite{chen2018isolating} are identical to VAEs, except that an additional penalty is applied to the total correlation between representation dimensions: \[
\mathcal{L}_{\mathrm{TC}}(x) = \mathcal{L}_{\mathrm{VAE}}(x)  + (\beta-1) \mathrm{TC}(\mathtt{enc}(x)),
\]where for some joint distribution $q(z)$, the total correlation $\mathrm{TC}(q(z))$ is equivalent to the KL divergence between $q(z)$ and the product of its marginals, $\mathrm{KL}(q(z)||\prod_j q(z_j))$. Penalizing total correlation reduces statistical dependence between dimensions and is thought to improve interpretability. We use an approximation of this objective developed by Chen et al. \cite{chen2018isolating} but there are others, e.g. Kim et al. \cite{kim2018disentangling}. For all experiments, we used $\beta=10$. 

Our semi-supervised (SS) variant of the $\beta$-total correlation autoencoder augments its encoded representation $z$ with an additional categorical dimension where we explicitly provide the class label $y$: \[
\begin{split}
	\mathcal{L}_{\mathrm{SS}}(x,y) &= \mathbb{E}_{z \sim \mathtt{enc}(x)} \left[ \mathcal{L}(x, \mathtt{concat}(y,\mathtt{dec}(z))) \right]\\
	&+ \mathrm{KL}(\mathtt{enc}(x)||p(z))\\
&+ (\beta-1) \mathrm{TC}(\mathtt{enc}(x))
\end{split}
\]
By providing $y$ as side-information during training, we effectively disentangle digit identity from the continuous part of the representation $z$, making the SS slightly more similar to ground-truth (even though there is no complete ground-truth for MNIST, the dataset we consider). The SS model's representation of digit style may still be entangled, but hopefully less so than other models (as the SS model still employs all the same tricks as the TC model).

Finally, InfoGAN \cite{chen2016infogan} (IG) is a generative adversarial network \cite{goodfellow2014generative} trained to reach equilibrium between two losses: a discriminator attempting to distinguish real from fake images, and generator attempting to create images $x$ that fool the discriminator from a latent code $z$ (with maximal information between ``interpretable'' components of $z$ and the generated image). We refer to the original citation for more details \cite{chen2016infogan}.

\subsubsection{Training Details.}

All autoencoder models are trained in Tensorflow with the Adam optimizer \cite{kingma2014adam}, with a batch size of 64 or 128 (for MNIST vs. others) and for the minimum number of epochs necessary to surpass 100,000 iterations. Matching Burgess et al.~\cite{burgess2018understanding}, the learning rate was set to 0.0005 for dSprites and its Tensorflow-default value of 0.001 for MNIST and Sinelines. InfoGANs were trained using an implementation from the Tensorpack library \cite{wu2016tensorpack}. See \url{https://github.com/dtak/interactive-reconstruction} for code.

\subsection{Distance Metrics and Thresholds}
\label{sec:metrics-thresholds}
In this section, we describe our choices of distance metric $d(x,x')$ and threshold distance $\epsilon$ for each of the three datasets. In general, for all three datasets, we used metrics that measured the fraction of disagreeing dimensions, with dataset-specific definitions of disagreement.

\subsubsection{Sinelines}
For Sinelines, we defined distance as the fraction of inputs that disagreed by more than 0.5 (approximately 2\% of the range of $x$): \begin{equation}
    d_{\mathrm{Sinelines}}(x,x^*) \triangleq \frac{1}{64} \sum_{i=1}^{64} \mathbbm{1}(|x_i-x^*_i| > 0.5)    
\end{equation}
This metric has the advantage of being between 0 and 1, which allows us to visualize users' proximity to solving each problem with a progress bar (after subtracting the distance from 1). We set the threshold distance value $\epsilon$ to 0.1 (which we visualized as a 90\% agreement target). 

\subsubsection{dSprites}
For dSprites, a BW image dataset where a large fraction of pixels are always within almost any threshold value due to black backgrounds, we instead used L1 distance normalized by the total difference away from black backgrounds (which corresponds to Bray-Curtis similarity \cite{beals1984bray}, and can be seen as a relaxation of intersection-over-union): \begin{equation}
  d_{\mathrm{dSprites}}(x,x^*) \triangleq \frac{\sum_i |x_i - x^*_i| }{ \sum_i |x_i| + |x^*_i| }
\end{equation}
We again used the threshold of $\epsilon=0.1$ and visualized progress in terms of a 90\% alignment target.

\subsubsection{MNIST}
For MNIST, we initially tried the same distance metric as with dSprites, but found in pilot experiments that the soft relaxation of intersection-over-union led to confusing behavior with highly regularized architectures like the $\beta$-total correlation autoencoder, which often have relatively gradual transitions between black and white portions of generated images. Instead, we opted to binarize autoencoder outputs in our visualization and use an exact intersection-over-union similarity metric: \begin{equation}
 d_{\mathrm{MNIST}}(x,x') \triangleq 1 - \frac{ \sum_{i} \round{x_i} \wedge \round{x_i'} }{ \sum_{i} \round{x_i} \lor \round{x_i'} },
\end{equation}
with a threshold of $\epsilon=0.25$ (visually presented as an alignment target of 75\%). We chose this threshold by asking pilot users to align $x$ and $x'$ without a target threshold and verbally indicating when they felt they were ``far away'' vs. ``close enough,'' and then finding a value that consistently separated those two states across examples.

\subsection{Full Tables of Results}
\label{sec:full-results}

\begin{figure*}
    \centering
    \includegraphics[width=\textwidth]{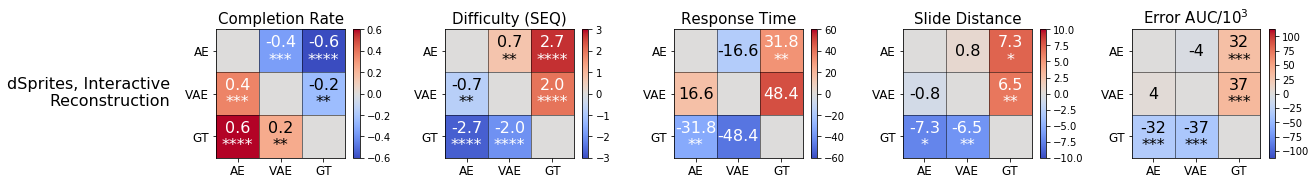}
    \includegraphics[width=\textwidth]{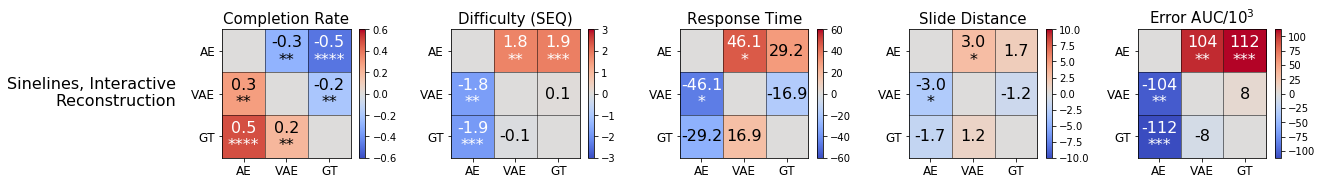}
    \includegraphics[width=\textwidth]{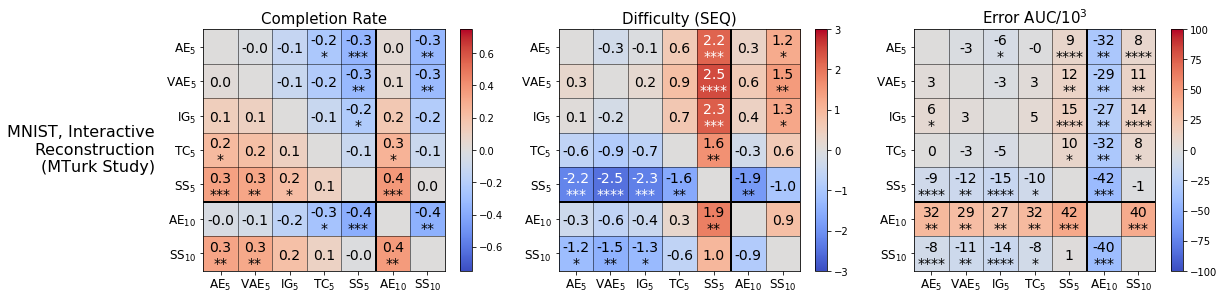}
    \includegraphics[width=\textwidth]{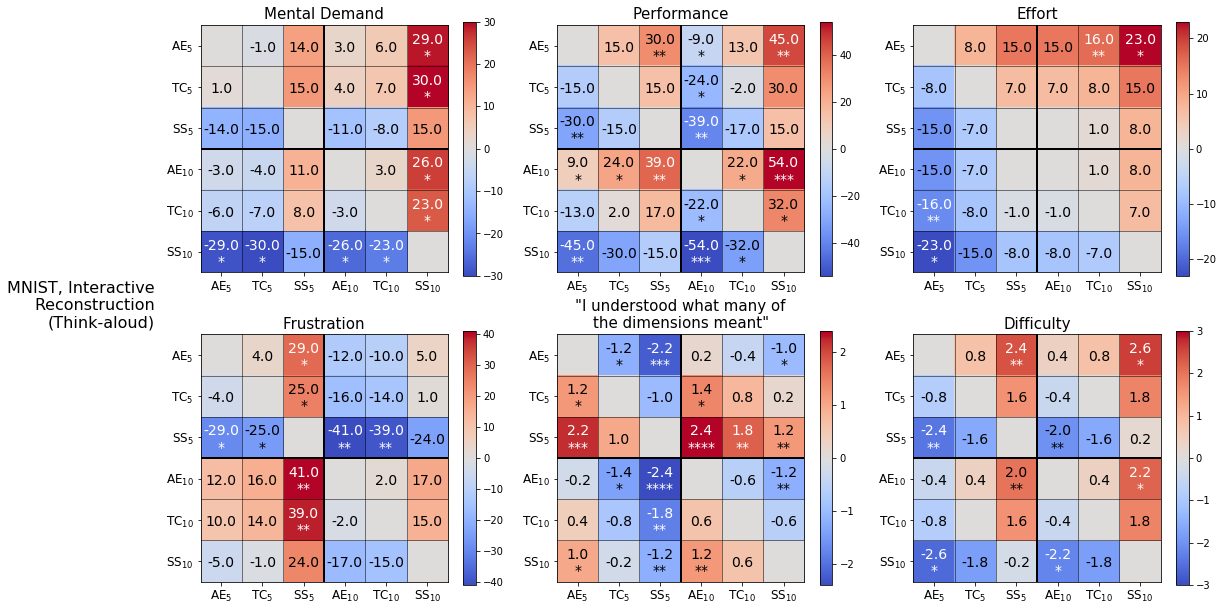}
    \caption{Pairwise average differences in interactive reconstruction metrics between models (row minus column). Stars indicate t-test p-values (****=p<0.0001, ***=p<0.001, **=p<0.01, *=p<0.05, paired for dSprites and Sinelines and unpaired for MNIST). Differences were largest and most significant between autoencoders (AE) and disentangled ground-truth (GT) or semi-supervised (SS) models (bottom-left or top-right corners in each plot or dark-line-delineated subgroup).}
    \label{fig:fr-pairwise}
\end{figure*}

\begin{figure*}
    \centering
    \includegraphics[width=0.925\textwidth]{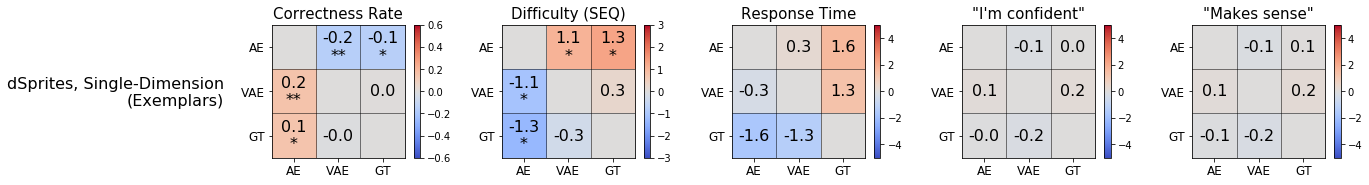}
    \includegraphics[width=0.925\textwidth]{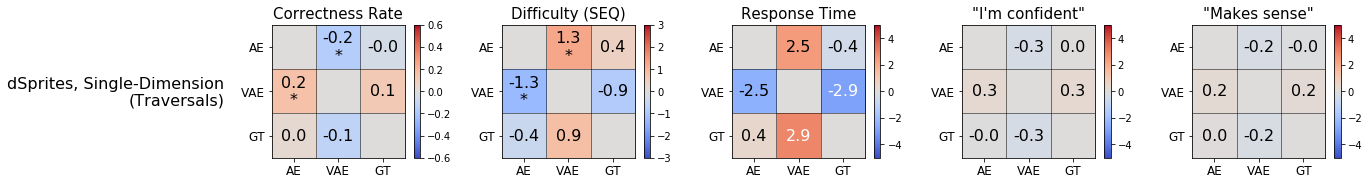}
    \includegraphics[width=0.925\textwidth]{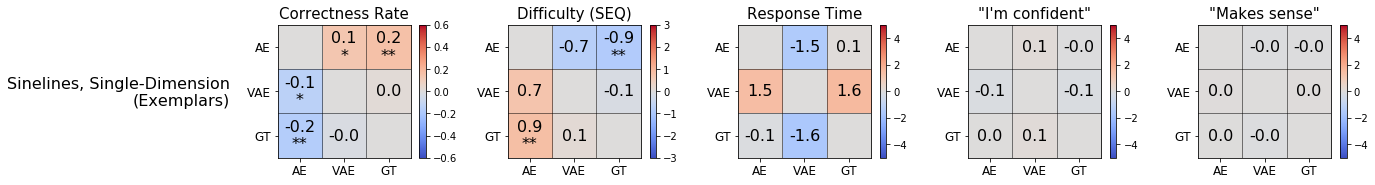}
    \includegraphics[width=0.925\textwidth]{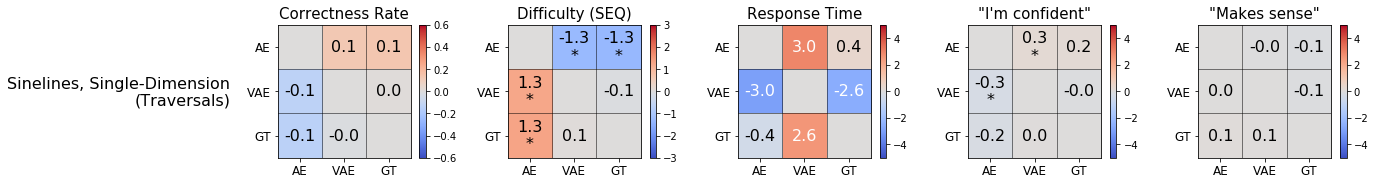}
    \caption{Pairwise differences in single-dimension metrics between models (rows, columns, and stars defined in Figure~\ref{fig:fr-pairwise}). Differences were smaller, less consistent with ground-truth disentanglement, and further from significance compared to interactive reconstruction.}
    \label{fig:sd-pairwise}
\end{figure*}

As an alternative to the plots in the main paper, we present results in tabular form in Table~\ref{tbl:synth-all} as well as heatmaps of pairwise differences in Figures~\ref{fig:fr-pairwise} and \ref{fig:sd-pairwise}. Note that synthetic result tables include a redundant post-task measure of difficulty, which we opted not to include (though it gives similar results) due to comments during the think-aloud study and from MTurk users that it was easy to forget the first model by the end of the quiz.

\subsection{Additional Screenshots}
\label{sec:extra-screenshots}

\begin{figure*}
    \centering
    \fbox{\includegraphics[width=0.25\linewidth]{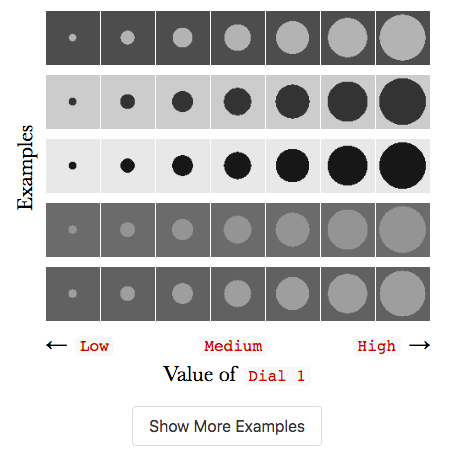}}
    \fbox{\includegraphics[width=0.25\linewidth]{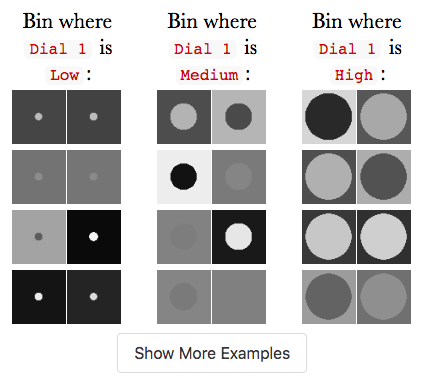}}
    \fbox{\includegraphics[width=0.4\linewidth]{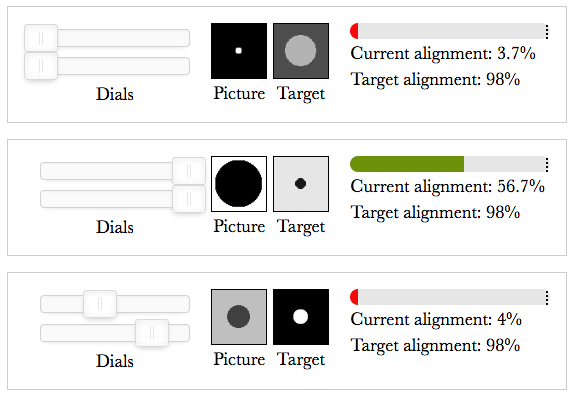}}
    \caption{Traversals (left), exemplars (middle), and interactive reconstruction (right) for practice question problem.}
    \label{fig:pqs}
\end{figure*}

\begin{figure*}
    \centering
        \fbox{\includegraphics[width=0.3\linewidth]{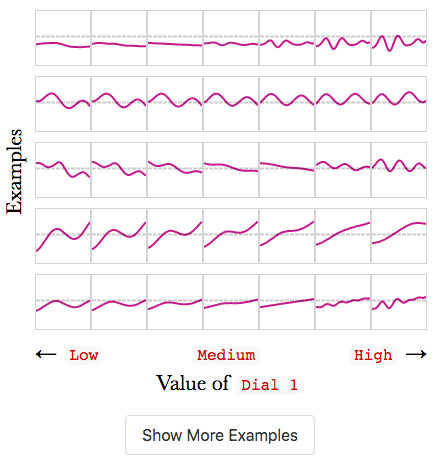}}
    \fbox{\includegraphics[width=0.3\linewidth]{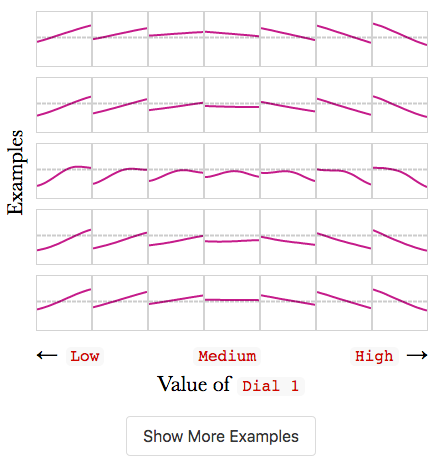}}
        \fbox{\includegraphics[width=0.3\linewidth]{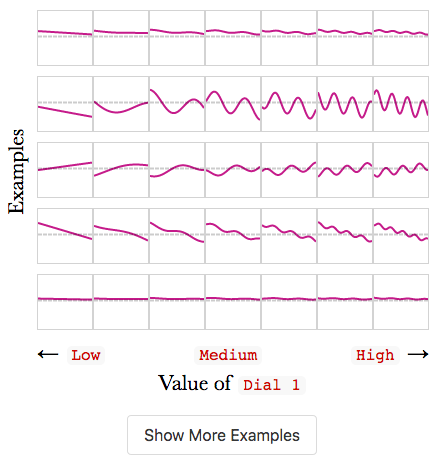}}
    \caption{Sinelines traversals for a random dimension of the AE (left), VAE (middle), and GT model (right).}
    \label{fig:traversals-sinelines}
\end{figure*}

\begin{figure*}
    \centering
        \fbox{\includegraphics[width=0.3\linewidth]{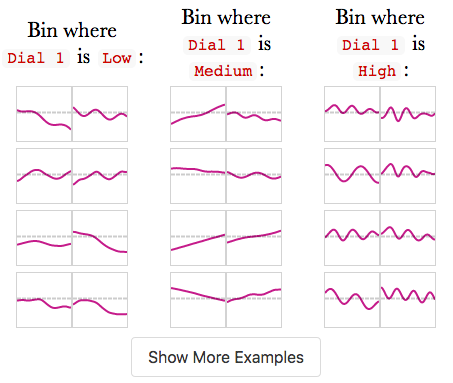}}
    \fbox{\includegraphics[width=0.3\linewidth]{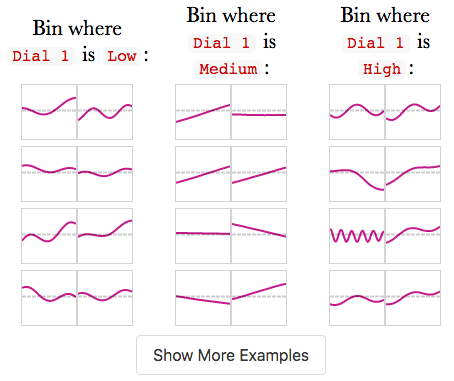}}
        \fbox{\includegraphics[width=0.3\linewidth]{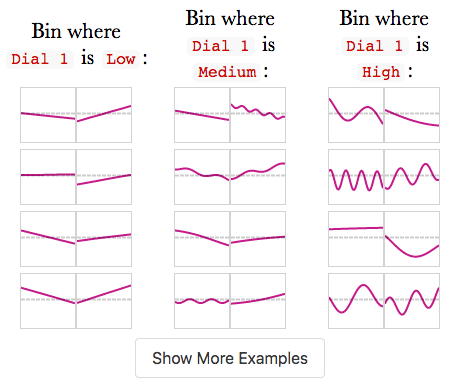}}
    \caption{Sinelines exemplars for a random dimension of the AE (top), VAE (middle), and GT model (bottom).}
    \label{fig:exemplars-sinelines}
\end{figure*}

In Figures \ref{fig:pqs}, \ref{fig:traversals-sinelines}, and \ref{fig:exemplars-sinelines}, we show additional screenshots for more tasks, models, and datasets. We also showcase the instructional practice questions, where users were shown a simple ``circles'' dataset with just two factors of variation (radius and foreground/background color). Note that all of the tasks are available at \url{http://hreps.s3.amazonaws.com/quiz/manifest.html}.

\begin{table*}
\centering
\begin{tabular}{llllllll}
\toprule
            dSprites Inter. Recon. &         AE &         VAE &         GT &  AnovaRM &    AE×VAE &     AE×GT &    VAE×GT \\
\midrule
       Difficulty (SEQ, 1-7) &  6.93±0.25 &   6.20±0.91 &  4.20±1.60 &  p=7e-08 &   p=0.006 &   p=2e-05 &   p=8e-05 \\
 Difficulty (Post-Quiz, 1-5) &  4.07±1.29 &   4.07±0.85 &  2.67±1.14 &  p=0.004 &     p=1.0 &    p=0.03 &   p=0.002 \\
            Error AUC / 1000 &  55.8±25.6 &   60.1±26.6 &   23.6±5.9 &  p=5e-06 &     p=0.3 &  p=0.0003 &  p=0.0002 \\
              Slide Distance &  13.6±10.1 &    12.8±8.4 &  6.24±2.28 &  p=0.003 &     p=0.7 &    p=0.01 &   p=0.006 \\
               Response Time &  80.6±47.2 &  97.2±103.0 &  48.8±23.4 &   p=0.08 &     p=0.5 &   p=0.006 &    p=0.07 \\
             Completion Rate &  0.15±0.19 &   0.51±0.33 &  0.75±0.26 &  p=5e-08 &  p=0.0009 &   p=6e-07 &   p=0.003 \\
\bottomrule
\end{tabular}
\begin{tabular}{llllllll}
\toprule
           Sinelines Inter. Recon. &          AE &        VAE &         GT &   AnovaRM &   AE×VAE &     AE×GT &   VAE×GT \\
\midrule
       Difficulty (SEQ, 1-7) &   5.87±1.67 &  4.07±1.84 &  4.00±1.71 &  p=0.0002 &  p=0.002 &  p=0.0007 &    p=0.9 \\
 Difficulty (Post-Quiz, 1-5) &   3.93±1.12 &  3.07±1.34 &  2.67±1.25 &   p=0.003 &   p=0.04 &   p=0.001 &    p=0.2 \\
            Error AUC / 1000 &  163.5±90.0 &  59.6±47.8 &  52.0±48.9 &   p=6e-05 &  p=0.002 &  p=0.0007 &    p=0.6 \\
              Slide Distance &   7.52±3.43 &  4.56±2.32 &  5.77±3.28 &    p=0.05 &   p=0.03 &     p=0.2 &    p=0.2 \\
               Response Time &   99.0±62.2 &  52.9±35.6 &  69.8±38.6 &    p=0.03 &   p=0.02 &     p=0.1 &    p=0.1 \\
             Completion Rate &   0.34±0.26 &  0.63±0.28 &  0.83±0.19 &   p=1e-06 &  p=0.002 &   p=2e-05 &  p=0.006 \\
\bottomrule
\end{tabular}

\begin{tabular}{llllllll}
\toprule
          dSprites SD (Exemplars) &         AE &        VAE &         GT &  AnovaRM &   AE×VAE &   AE×GT &  VAE×GT \\
\midrule
       Difficulty (SEQ, 1-7) &  5.33±1.25 &  4.27±1.18 &  4.00±1.63 &   p=0.03 &   p=0.02 &  p=0.03 &   p=0.6 \\
 Difficulty (Post-Quiz, 1-5) &  3.13±1.09 &  2.27±0.77 &  2.87±0.96 &   p=0.06 &   p=0.05 &   p=0.4 &  p=0.08 \\
               Response Time &   16.9±6.4 &   16.6±8.6 &   15.3±7.5 &    p=0.8 &    p=0.9 &   p=0.4 &   p=0.7 \\
     ``I'm confident'' (1-5) &  3.99±0.67 &  4.12±0.48 &  3.95±0.70 &    p=0.3 &    p=0.2 &   p=0.8 &  p=0.06 \\
       ``Makes sense'' (1-5) &  3.86±0.73 &  4.01±0.55 &  3.80±0.95 &    p=0.3 &    p=0.1 &   p=0.7 &   p=0.2 \\
            Correctness Rate &  0.61±0.18 &  0.76±0.14 &  0.75±0.14 &  p=0.008 &  p=0.009 &  p=0.01 &   p=0.9 \\
\bottomrule
\end{tabular}
\begin{tabular}{llllllll}
\toprule
         Sinelines SD (exemplars) &         AE &        VAE &         GT &  AnovaRM &  AE×VAE &    AE×GT & VAE×GT \\
\midrule
       Difficulty (SEQ, 1-7) &  4.47±1.59 &  5.20±1.42 &  5.33±1.49 &   p=0.04 &  p=0.09 &  p=0.003 &  p=0.7 \\
 Difficulty (Post-Quiz, 1-5) &  2.67±1.19 &  3.00±1.21 &  3.27±1.24 &    p=0.1 &   p=0.3 &   p=0.03 &  p=0.3 \\
               Response Time &   12.5±4.8 &   14.0±5.5 &   12.4±4.9 &    p=0.5 &   p=0.3 &    p=1.0 &  p=0.4 \\
     ``I'm confident'' (1-5) &  3.87±0.79 &  3.75±0.83 &  3.89±0.75 &    p=0.5 &   p=0.5 &    p=0.9 &  p=0.2 \\
       ``Makes sense'' (1-5) &  3.73±0.95 &  3.77±0.84 &  3.73±0.84 &    p=1.0 &   p=0.8 &    p=1.0 &  p=0.8 \\
            Correctness Rate &  0.81±0.15 &  0.67±0.11 &  0.65±0.11 &  p=0.005 &  p=0.02 &  p=0.003 &  p=0.7 \\
\bottomrule
\end{tabular}
\begin{tabular}{llllllll}
\toprule
         dSprites SD (traversals) &         AE &        VAE &         GT & AnovaRM &  AE×VAE &  AE×GT &  VAE×GT \\
\midrule
       Difficulty (SEQ, 1-7) &  4.87±1.36 &  3.60±1.31 &  4.47±1.26 &  p=0.03 &  p=0.01 &  p=0.2 &   p=0.1 \\
 Difficulty (Post-Quiz, 1-5) &  2.93±1.24 &  2.53±1.09 &  3.07±1.06 &   p=0.3 &   p=0.4 &  p=0.5 &   p=0.2 \\
               Response Time &  16.8±11.7 &   14.3±7.9 &  17.2±12.9 &   p=0.5 &   p=0.3 &  p=0.9 &   p=0.3 \\
     ``I'm confident'' (1-5) &  3.45±0.87 &  3.74±0.47 &  3.44±0.66 &   p=0.2 &   p=0.2 &  p=1.0 &  p=0.07 \\
       ``Makes sense'' (1-5) &  3.37±0.93 &  3.61±0.58 &  3.41±0.79 &   p=0.4 &   p=0.2 &  p=0.9 &   p=0.2 \\
            Correctness Rate &  0.56±0.19 &  0.73±0.20 &  0.60±0.19 &  p=0.03 &  p=0.01 &  p=0.5 &  p=0.08 \\
\bottomrule
\end{tabular}
\begin{tabular}{llllllll}
\toprule
        Sinelines SD (traversals) &         AE &        VAE &         GT & AnovaRM &  AE×VAE &   AE×GT & VAE×GT \\
\midrule
       Difficulty (SEQ, 1-7) &  3.67±1.78 &  4.93±1.69 &  5.00±1.15 &  p=0.01 &  p=0.04 &  p=0.03 &  p=0.8 \\
 Difficulty (Post-Quiz, 1-5) &  2.53±1.15 &  2.87±0.96 &  2.87±1.20 &   p=0.5 &   p=0.3 &   p=0.4 &  p=1.0 \\
               Response Time &  18.8±12.4 &   15.8±7.1 &   18.4±7.0 &   p=0.4 &   p=0.3 &   p=0.9 &  p=0.1 \\
     ``I'm confident'' (1-5) &  3.95±0.73 &  3.69±0.76 &  3.73±0.67 &  p=0.06 &  p=0.05 &  p=0.09 &  p=0.7 \\
       ``Makes sense'' (1-5) &  3.74±0.98 &  3.77±0.61 &  3.83±0.66 &   p=0.8 &   p=0.9 &   p=0.5 &  p=0.5 \\
            Correctness Rate &  0.71±0.18 &  0.59±0.17 &  0.57±0.14 &  p=0.08 &   p=0.1 &  p=0.08 &  p=0.8 \\
\bottomrule
\end{tabular}
\caption{Full tables of results for the synthetic dataset MTurk study (as an alternative view to the boxplots in Figure \ref{fig:boxplots} of the main paper). Values on the denote means and standard deviations. p-values are for ANOVA with repeated measures (middle) and paired t-tests (right). Results include a redundant post-task measure of difficulty, which we opted not to include (though it gives similar results) due to comments during the qualitative evaluation that it was easy to forget the first model by the end of the quiz.}
\label{tbl:synth-all}
\end{table*}

\begin{table*}[h!]
    \centering
    \small
\begin{tabular}{|c|c|l|}
\hline
Model &      Dimension &                                      User-Assigned Labels \\
\hline\hline
 \multirow{3}{*}{$\mathrm{AE}_{10}$} &  Continuous 1 &  \texttt{"cross"} \\\cline{2-3}
  &  Continuous 4 &  \texttt{"intensity", "0-1"} \\\cline{2-3}
  &  Continuous 7 &  \texttt{"thickness"} \\\hline\hline
 \multirow{10}{*}{$\mathrm{TC}_{10}$} &  Continuous 1 &  \texttt{"numbers", "reflect but not really", "number?", "little"} \\\cline{2-3}
  &  Continuous 2 &  \texttt{\textcolor{gray}{"x", "ignore", "nothing", "no"}} \\\cline{2-3}
  &  Continuous 3 &  \texttt{"thickness", "numb"} \\\cline{2-3}
  &  Continuous 4 &  \texttt{"expands", "numb"} \\\cline{2-3}
  &  Continuous 5 &  \texttt{\textbf{"makes things thick", "thickness", "thicker"}} \\\cline{2-3}
  &  Continuous 6 &  \texttt{\textcolor{gray}{"x", "ignore", "nothing", "nothing", "no"}} \\\cline{2-3}
  &  Continuous 7 &  \texttt{\textcolor{gray}{"x", "ignore", "nothing", "nothing", "no"}} \\\cline{2-3}
  &  Continuous 8 &  \texttt{\textbf{"tilt on center", "rotation", "orient"}} \\\cline{2-3}
  &  Continuous 9 &  \texttt{"squish into the center", "numb"} \\\cline{2-3}
  &  Continuous 10 & \texttt{"six", "stretch", "0"} \\\hline\hline
 \multirow{11}{*}{$\mathrm{SS}_{10}$} &  Discrete 1 &  \texttt{\textbf{"Number", "Number", "number"}} \\\cline{2-3}
 &  Continuous 1 &  \texttt{"spirals", "lengthening", "width"} \\\cline{2-3}
 &  Continuous 2 &  \texttt{"curviness", "y bubble"} \\\cline{2-3}
 &  Continuous 3 &  \texttt{\textcolor{gray}{"x", "ignore", "nothing", "no"}} \\\cline{2-3}
 &  Continuous 4 &  \texttt{\textbf{"thickness", "thick", "thickness", "thick", "thickness"}} \\\cline{2-3}
 &  Continuous 5 &  \texttt{"rotation"} \\\cline{2-3}
 &  Continuous 6 &  \texttt{\textcolor{gray}{"x", "ignore", "nothing", "little"}} \\\cline{2-3}
 &  Continuous 7 &  \texttt{\textcolor{gray}{"x"}, "moves a thing", \textcolor{gray}{"no"}} \\\cline{2-3}
 &  Continuous 8 &  \texttt{"tilt", "squishes"} \\\cline{2-3}
 &  Continuous 9 &  \texttt{\textbf{"semi tilt", "rotation"}} \\\cline{2-3}
 &  Continuous 10 & \texttt{\textcolor{gray}{"x", "migrates one thing", "nothing", "little"}} \\\hline

\end{tabular}
    \caption{All labels assigned to $D_z=10$ models by participants in the MNIST thinkaloud study. Bold font shows labels experimenters identified as consistent between participants, with dimensions that had little effect on representations due to TC regularization shown in gray. TC and SS models had more labels (and more consistent labels) than AEs.}
    \label{fig:mnist-labeling-10d}
\end{table*}

\end{document}